\documentclass[a4paper,fleqn]{cas-dc}

\usepackage[authoryear]{natbib}
\usepackage{makecell}
\usepackage{caption}
\usepackage{graphicx}
\usepackage{booktabs}
\usepackage{array}
\usepackage{verbatim}
\usepackage{ulem}
\definecolor{light-gray}{gray}{0.95}

\def\tsc#1{\csdef{#1}{\textsc{\lowercase{#1}}\xspace}}
\tsc{WGM}
\tsc{QE}%每当在文档中使用\WGM或\QE时，它们就会以小型大写字母格式显示，并且它们的文本将是小写转为的小型大写（例如，\WGM显示为 Wgm，而\QE显示为 Qe）
\newproof{pf}{Proof}
\newproof{pot}{Proof of Theorem \ref{thm}}
\begin{document}
	\let\WriteBookmarks\relax
	\def\floatpagepagefraction{1}
	\def\textpagefraction{.001}
	
	\shorttitle{PH2ST}    
	
	\shortauthors{Y. Niu et al.}
	
	\title [mode = title]{{PH2ST}: Prompt-Guided Hypergraph Learning for Spatial Transcriptomics Prediction in Whole Slide Images}
	
	%\tnotemark[<tnote number>] 
	
	%\tnotetext[<tnote number>]{<tnote text>} #标题的页脚注释
	
	\author[1]{Yi Niu}
	%\ead{dzhang@xjtu.stu.edu.cn}
	\author[1]{Jiashuai Liu}
	
    \author[1]{Yingkang Zhan}
    \author[1]{Jiangbo Shi}
    \author[1]{Di Zhang}
    \author[2,3,4,5]{Marika Reinius}
    \author[2,3]{Ines Machado}
    \author[2,3]{Mireia Crispin-Ortuzar}
    \author[6]{Jialun Wu}
	%\credit{Conceptualization of this study, Methodology, Software}
	%\author[<aff no>]{<author name>}[<options>]
    \author[1]{Chen Li} \cormark[1] \ead{cli@xjtu.edu.cn}
    \author[2,3]{Zeyu Gao} \cormark[1] \ead{zg323@cam.ac.uk}
	%\credit{supervision, writing}
    \cortext[1]{Co-Corresponding author.}
	
	%\fntext[2]{Ming Da is a professor in Sky University. He is interested in ... ...}
	%\cormark[<corr mark no>]
	
	%\fnmark[<footnote mark no>]
	

	%\ead[url]{<URL>}
	

\affiliation[1]{organization={School of Computer Science and Technology, Xi'an Jiaotong University},
city={Xi'an},
postcode={710049}, 
country={China}}
\affiliation[2]{organization={Department of Oncology, University of Cambridge},
			city={Cambridge},
			postcode={CB2 0XZ}, 
			country={UK}}
\affiliation[3]{organization={CRUK Cambridge Centre, University of Cambridge},
			city={Cambridge},
			postcode={CB2 0RE}, 
			country={UK}}
\affiliation[4]{organization={CRUK Cambridge Institute, University of Cambridge},
			city={Cambridge},
			postcode={CB2 0RE}, 
			country={UK}}
\affiliation[5]{organization={Cambridge University Hospitals NHS Foundation Trust},
			city={Cambridge},
			postcode={CB2 0QQ}, 
			country={UK}}
\affiliation[6]{organization={School of Computer Science, Northwestern Polytechnical University},
  city={Xi'an},
  postcode={710072},
  country={China}}
	
	% Email id of the second author
	%\ead{}



	
	
	\begin{abstract}
        Spatial Transcriptomics (ST) reveals the spatial distribution of gene expression in tissues, offering critical insights into biological processes and disease mechanisms. 
        However, the high cost, limited coverage, and technical complexity of current ST technologies restrict their widespread use in clinical and research settings, making obtaining high-resolution transcriptomic profiles across large tissue areas challenging.
        Predicting ST from H\&E-stained histology images has emerged as a promising alternative to address these limitations but remains challenging due to the heterogeneous relationship between histomorphology and gene expression, which is affected by substantial variability across patients and tissue sections.
        In response, we propose PH2ST, a prompt-guided hypergraph learning framework, which leverages limited ST signals to guide multi-scale histological representation learning for accurate and robust spatial gene expression prediction. 
        Extensive evaluations on two public ST datasets and multiple prompt sampling strategies simulating real-world scenarios demonstrate that PH2ST not only outperforms existing state-of-the-art methods, but also shows strong potential for practical applications such as imputing missing spots, ST super-resolution, and local-to-global prediction, highlighting its value for scalable and cost-effective spatial gene expression mapping in biomedical contexts.
        % Benchmark evaluations on two public ST datasets demonstrate that PH2ST outperforms the existing state-of-the-art methods and closely aligns with the ground truth. 
        % Furthermore, we evaluate PH2ST under multiple prompt sampling strategies that simulate real-world scenarios, demonstrating its potential for applications such as imputing missing discrete spots, ST super-resolution, and local-to-global prediction.
        % These results underscore the potential of leveraging limited ST data for scalable and cost-effective spatial gene expression mapping in real-world biomedical applications.

	\end{abstract}
	%Enhancing the reliability of models in computational pathology is therefore of utmost importance.
	
	\begin{highlights}
    % \item \textcolor{red}{\textbf{TOO LONG!!! AND SHOULD NOT INCLUDE IN MANUSCRIPT!!! Highlights are required for original articles. They consist of a small collection of topics (bullets,markers) that summarize the main findings of the article. They must be sent in an editable file, including 3 to 5 topics (maximum 85 characters including spaces, per topic}).}
    % \item We redefine the spatial transcriptomics (ST) prediction task to better reflect clinical practice by introducing an inference-time prompting paradigm, where a small subset of known spots per WSI guides the prediction of gene expression in the remaining regions.
    
    % \item We propose PH2ST, a novel framework that leverages limited ST data as prompts to guide the learning of associations between histological features and gene expression, improving spatial prediction performance.
    
    % \item PH2ST integrates multi-scale histological features and constructs hypergraphs to fuse spatial and contextual information from tissue morphology effectively.
    
    % \item We evaluate PH2ST on two publicly available ST datasets under four prompting scenarios aligned with real-world needs, demonstrating state-of-the-art performance and strong practical applicability.
    
    \item {Redefined spatial transcriptomics~(ST) prediction as an inference-time prompting task.}

    \item {A novel framework using limited ST prompts to guide spatial gene expression prediction.}

    \item {Effective dual-scale hypergraph for integrating multi-scale histology features and spatial context.}

    \item {State-of-the-art performance and strong generalization in real-world ST tasks.}

    % \item \textcolor{green}{Demonstrated strong applicability in four real-world scenarios.}
    \end{highlights}
	
	% Keywords
	
	\maketitle
	
	% Main tex
	\section{Introduction}\label{introduction}

% \textcolor{red}{\textbf{\textit{GRAPHIC INTRODUCTION REQUIRED! It is recommended to include a concise and illustrative figure in the Introduction section. This figure could serve to visually convey the problem addressed by the work, highlight the key challenges or limitations of existing approaches, and provide an overview of the strategies adopted in the proposed method. Such a schematic would significantly enhance clarity for both reviewers and readers}}.}
    
    	% \textcolor{red}{As the gold standard for pathological diagnosis of cancer, whole slide image (WSI) provides an important reference for pathologists' clinical diagnosis-making \citep{wu2022recent, gao2020renal}. Genetic testing is of great significance for cancer targeted therapy and assessment of recurrence and metastasis risk \citep{ash2021joint}. At present, the technology has been able to measure gene expression at single-cell resolution even provide spatial information, laying a foundation for dissecting the relationship between tissue phenotype and gene expression \citep{alfonzo2021call}.}
        Histopathological slides remain the gold standard for cancer diagnosis, providing crucial morphological information that guides pathologists in clinical decision-making. Morphology-based diagnosis enables the identification of tumor types, grades, and invasion patterns \citep{amin2017ajcc,cheung2018classification}.
        However, with the growing understanding of cancer biology, it has become increasingly evident that morphological features alone are insufficient for accurate diagnosis and personalized treatment \citep{10002010map,barabasi2004network,ober2011gene}.  Many diagnostic and prognostic decisions now rely not only on histological morphology but also on molecular profiling, including genetic mutations, gene expressions, and DNA methylation status \citep{garraway2013lessons,schnitt2010classification,papanicolau2022dna}.

        Traditional bulk sequencing averages gene expression across heterogeneous cell populations \citep{wang2009rna}, obscuring spatial context and making it difficult for researchers and oncologists to fully understand tumor heterogeneity and its microenvironment. 
        This limits insights into cancer progression and hinders the development of effective new treatments.
        In response to the limitations of traditional sequencing, spatial transcriptomics (ST) has emerged as a transformative technology that enables the spatially resolved quantification of gene expression within tissue sections \citep{staahl2016visualization,cui2022spatially,tian2023expanding}. By integrating transcriptomic data with histological context at near-cellular resolution, ST allows researchers to link molecular profiles with tissue morphology, providing a powerful tool for investigating tumor heterogeneity, cellular interactions, and spatially organized signaling pathways \citep{alfonzo2021call,moncada2020integrating,rao2021exploring}.
        Despite its promise, however, the large-scale application of ST remains limited by high costs, technical complexity, and labor-intensive protocols \citep{palla2022spatial}.

        Given these constraints and the transformative potential of ST, computational pathology models that infer spatial transcriptomic profiles from routinely available H\&E-stained whole slide images (WSIs) are being actively explored as a promising direction \citep{lee2022deep}.
        While a growing number of studies have shown promising results in predicting WSI-level molecular characteristics, such as biomarkers \citep{el2024regression}, DNA mutations \citep{yan2023histopathological}, RNA expression \citep{pizurica2024digital}, and DNA methylation \citep{hoang2024prediction}, from WSIs using weakly supervised learning based models, the prediction of spatial transcriptomics remains in its early stages due to data scarcity and technical challenges.
        Several recent works have begun to explore this direction. 
        ST-Net \citep{he2020integrating}, as one of the initial efforts, uses individual spot images with a DenseNet backbone for local expression prediction.
        To capture broader spatial dependencies, HisToGene \citep{pang2021leveraging} and Hist2ST \citep{zeng2022spatial} incorporate Transformer and Graph Neural Network (GNN) architectures, while THItoGene \citep{jia2024thitogene} and HGGEP \citep{li2024gene} further integrate global and local spot representations through stacked Vision Transformers and GNNs.
        In addition, BLEEP \citep{xie2024spatially} employs contrastive learning to align spatial gene expression with histological features, and TRIPLEX \citep{chung2024accurate} highlights the importance of multi-scale feature extraction for improving ST prediction.

        While existing methods have demonstrated the feasibility of predicting ST from histological images, three fundamental and critical limitations remain unresolved.
        (1) \textit{Mismatch with Clinical Needs.}
        Existing ST prediction task settings remain misaligned with practical clinical demands \citep{jain2024spatial}. Due to technical constraints and high costs, widely used ST profiling technologies (\textit{e.g.}, 10x Visium) \citep{janesick2022high} are limited to small tissue regions, lack sufficient resolution for genuine single-cell analysis, and frequently suffer from significant dropout rates, deteriorating data quality \citep{kharchenko2014bayesian, lopez2019joint}. Consequently, practical ST prediction scenarios should shift toward imputing missing gene expressions, inferring gene expression in unmeasured areas based on limited local ST measurements, and enhancing ST super-resolution. 
        (2) \textit{Multi-Scale Modeling Deficiency.}
        Current methods inadequately model the complex and multi-scale spatial relationships between tissue phenotypes and gene expression \citep{burgess2019spatial}, struggling particularly to capture hierarchical spatial structures and long-range dependencies within WSIs. 
        (3) \textit{Domain Shift Vulnerability.}
        Significant variability across patients and tissue sections leads to substantial domain shifts, a problem exacerbated by limited scale and diversity of existing ST datasets \citep{jaume2024hest, zhang2022clinical}.

        To comprehensively overcome these limitations, we propose PH2ST—a novel \textbf{P}rompt-guided \textbf{H}ypergraph learning framework \textbf{to} predict \textbf{S}patial \textbf{T}ranscriptomics in WSIs, as shown in Fig. \ref{intro_graph}. Specifically, PH2ST addresses the above limitations through the following targeted solutions:
        (1) \textit{To address the mismatch with real clinical needs}, PH2ST introduces a novel inference-time prompting paradigm. 
        Instead of relying solely on paired WSI-ST training datasets, it leverages a limited number of spots with known ST measurements to guide inference.
        This enables accurate prediction of gene expression across unmeasured regions within each WSI,
        % Under this setting, ST profiles from a small subset of known spots within each WSI guide the inference of gene expression across the remaining unmeasured regions, thus 
        aligning prediction tasks closely with real clinical requirements, \textit{e.g.}, ST imputation, local-to-global prediction, and super-resolution.
        (2) \textit{To improve the modeling of complex, multi-scale spatial relationships}, PH2ST proposes a dual-scale hypergraph learning module. At the global scale, slide-level spot hypergraphs aggregate features across entire WSIs, capturing long-range dependencies among distant tissue regions. At the local scale, hyper-subgraphs integrate contextual spatial information from neighboring spots, enabling hierarchical modeling of fine-grained tissue structures and phenotypes.
        (3) \textit{To mitigate the domain shift caused by patient and tissue variability}, PH2ST incorporates a cross-attention mechanism to enable ST prompt-guided representation refinement. This component explicitly aligns local spot representations with ST prompts obtained from the same WSI, effectively reducing the impact of limited training data and enhancing model robustness and generalization to unseen slides.
        Experimental results from two benchmark datasets demonstrate that PH2ST consistently outperforms existing methods across all evaluation metrics. Further validation under various prompt sampling strategies, including fully random, sparse/dense local square-window, and evenly distributed Poisson disk sampling, confirms the robustness, accuracy, and scalability of PH2ST in realistic, resource-constrained scenarios.      
        The key contributions of this work are summarized as follows:

    \begin{figure}[t]
            \includegraphics[width=0.5\textwidth]{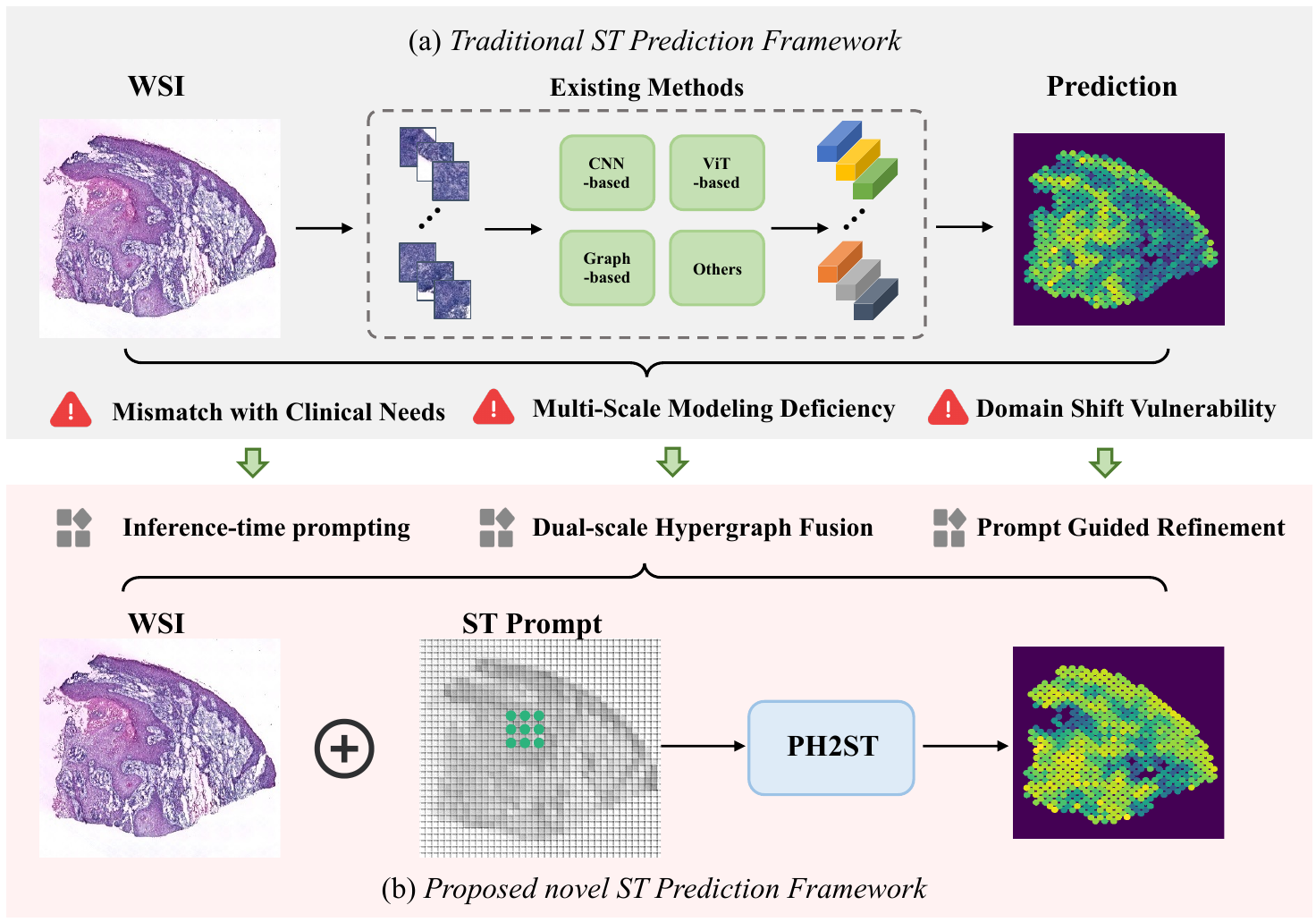}
            \caption{Illustration of (a) traditional ST prediction and (b) the proposed inference-time prompting paradigm.}
        \label{intro_graph}
    \end{figure}
        
        % \textcolor{red}{\textbf{\textit{The following CONTRIBUTION section presents several contributions that appear to overlap with the “solutions to limitations” outlined in the revised preceding paragraph. It may be beneficial to revisit and clearly summarize the specific contributions. For instance, consider explicitly stating points such as “To the best of our knowledge, this is the first work to integrate multimodal data in this manner,” or similar claims that highlight the novelty of the work}}.}
    	
    % \begin{itemize}
    %     % 加一个设定上的创新
    %     \item We introduce a new ST prediction paradigm that leverages a few spots with known gene expression to guide the inference of remaining regions, supporting diverse real-world scenarios in a cost-effective and clinically applicable manner.
        
    %     \item We propose PH2ST, a prompt-guided histological hypergraph learning framework that models both global and local spatial dependencies through dual-scale hypergraphs and effectively incorporates ST-prompt information via cross-attention mechanisms, thereby bridging histological features and spatial gene expression in the presence of limited prompt signals.
    %      % PH2ST integrates multi-scale histological features and constructs hypergraphs to fuse both spatial and contextual information from tissue images.
           
    %     \item	We validate PH2ST on two public ST datasets and across multiple simulated prompt usage scenarios, demonstrating superior performance over state-of-the-art methods and strong robustness under diverse ablation and generalization settings.
        
    % \end{itemize}
    \begin{itemize}
        \item {We propose a novel ST prediction framework} that leverages a few spatial spots with known gene expression as prompts to guide the inference of unmeasured regions, effectively enabling low-cost and clinically applicable ST in practical scenarios. To the best of our knowledge, this is the first work to frame ST prediction under a prompt-based framework.
    
        \item {We design PH2ST}, a prompt-guided hypergraph learning framework that integrates dual-scale histological hypergraphs to capture local and global spatial dependencies simultaneously. 
        The model further incorporates prompt signals via cross-attention to explicitly align histological representations with limited known spatial gene expressions.
    
        \item {We conduct extensive experiments on two public ST datasets}. PH2ST achieves state-of-the-art performance and demonstrates strong robustness and generalizability across diverse ablation studies and various simulated prompt usage scenarios.
    \end{itemize}

	% \begin{figure}[t]%[htbp]
	% 	\centering
	% 	\includegraphics[width=8.2cm]{assump.pdf} % 替换为你的图片文件名StaDis_assumption.png
	% 	\caption{Graphical illustration of the StaDis method. We hypothesize that the features of perturbed images deviate from the characteristics of the training set. As a result, OOD data shows a higher Stability distance.}
	% 	\label{f1}	
	% \end{figure}

\section{Related works}\label{related}
	%asdasdadsasdasd \citep{odin}
    \subsection{Graph-based WSI Analysis}
    In recent years, deep learning methods, especially those based on Convolutional Neural Networks (CNNs), have become the mainstream approach for digital pathology image analysis and computer-aided diagnosis \citep{lu2021data, li2021dual, shao2021transmil}. CNNs can effectively extract visual features from WSIs and apply them to downstream tasks. However, the limited capacity of CNNs to analyze complex, high-dimensional features restricts their potential in handling complex medical images \citep{zhou2020graph}.
    Recent years have witnessed the increasing application of GNNs in recommender systems and biomedicine \citep{wu2022graph, rao2021imputing, chen2021whole}, thanks to their ability to effectively model structural information. 
    GNNs, a type of deep learning model based on graph structures (nodes and edges), have shown great potential. Nodes represent entities, while edges represent relationships or connections between them. GNNs leverage edge connections to aggregate features between nodes, capturing hidden dependencies \citep{zhou2020graph}. This approach has proven particularly effective in downstream tasks related to pathological images. Representing cells, tissues, and WSI patches as nodes, and spatial and functional relationships as edges allows GNNs to effectively capture tissue characteristics for various analytical tasks \citep{ahmedt2021graph}.
    Graph learning has demonstrated great potential in molecular biology research, particularly in predicting biomarkers such as gene expression, methylation and characterizing the tumor microenvironment \citep{schapke2021epgat, liu2023rmdgcn, wang2021cell}, providing novel biological insights.
    Lately, a growing number of spatial transcriptomics prediction studies have adopted GNNs as their network architecture, achieving remarkable results. 
    Hist2ST combines CNNs, Transformers, and GNNs sequentially to predict spatial domain expression \citep{zeng2022spatial}. THItoGene incorporates a Graph Attention Network (GAT) for tissue-specific feature fusion, exploring the relationship between high-resolution pathological image phenotypes and gene expression regulation \citep{jia2024thitogene}. HGGEP explores higher-order correlations between multiple latent-stage features and utilizes a hypergraph network to establish connections between features at different scales \citep{li2024gene}. ErwaNet enhances gene prediction by capturing local interactions and global structural information within spot images, without requiring prior gene expression data \citep{2024Edge}. SpaGCN aggregates gene expression levels of each spot with its neighbors to identify spatial domains with consistent expression and histology \citep{hu2021spagcn}. While these graph-based methods offer modeling flexibility and utilize various topological relationships to model spot representations, they lack the incorporation of multi-scale tissue features into the GNN, missing a unified representation that combines spot image location information with multi-scale information.

	\subsection{Spatial Gene Expression Prediction}
	ST uses microarray technology to preserve tissue location information on a chip and uses second-generation sequencing to analyze mRNA within the tissue sample. By overlaying the sequencing data back onto the tissue image, an ST spot of the tissue section is generated \citep{zhang2022clinical}.
	
	The prediction task is modeled as a multi-output regression problem, using spot images as inputs and spot gene expression as labels. Previous works have explored various modules for this task. ST-Net uses DenseNet101 \citep{huang2017densely} to extract image embeddings for each spot and employs a linear layer for classification \citep{he2020integrating}. HisToGene \citep{pang2021leveraging} utilizes Vision Transformer \citep{alexey2020image} to learn relationships between spots. BLEEP \citep{xie2024spatially} introduces contrastive learning to predict the gene expression of query spots by retrieving similar gene expression profiles from the training set. Hist2ST \citep{zeng2022spatial} makes an attempt to graph neural networks. HGGEP \citep{li2024gene} uses a hypergraph among multiple latent stage features to form the global association. 
    Recently, TRIPLEX \citep{chung2024accurate} achieved promising results by fusing multi-scale histological images through cross-attention, highlighting the importance of spot and neighboring information for gene expression prediction. 
    Previous work has relied entirely on mapping spot images to ST data, often leading to suboptimal accuracy. In contrast, we incorporate ST prompts from limited regions to guide prediction, achieving more precise spatial gene expression mapping that better aligns with real-world scenarios.
	
	\subsection{Benchmark Setting for ST Prediction}
	
	In the ST prediction task, due to data scarcity, most studies \citep{pang2021leveraging,zeng2022spatial,li2024gene} employ the leave-one-out strategy. 
    Current ST datasets often contain multiple sections from the same tissue region of the same patient. This can lead to the performance being overestimated, as models are frequently trained and tested on the same patients. TRIPLEX \citep{chung2024accurate} emphasized the importance of cross-validation and established a new benchmark to evaluate previous methods. 
    Based on this benchmark, we introduce an inference-time prompting paradigm, where a small subset of spot expressions is known in each WSI during testing. Fine-tuning was applied on all comparison methods except BLEEP, using 10\% known ST spots per WSI from the test set. Further details are described in the experiment section.

% 	\begin{figure*}[t]%[htbp]
% 	\centering
% 	\includegraphics[width=\textwidth]{OOD_1.pdf} 
% 	\caption{StaDis framework. (a): A pathological image is segmented into tens of thousands of patches. (b): Double Feature Extractor module. Perturbation preprocessing is applied to the original image to generate a Perturbation image. Features are then extracted from both the original and Perturbation images. (c): Slide-level MIL Aggregation Module. The original and perturbation patch features are aggregated into a slide-level representation via the attention module, with both feature sets sharing the same attention module weights. (d): Patch/Slide-level ID Classification and OOD Detection Module. Notably, for patch-level OOD detection, Subfigure (c) is bypassed. In the diagram, black arrows represent model training and ID classification inference, while red arrows indicate the calculation of StaDis OOD score.}
% 	\label{fig:wide}	
% \end{figure*}

\section{Method}\label{method}
	% 这块简单介绍一下方法的总览

\subsection{Problem Formulation} \label{overview_PH2ST}

\begin{figure*}[t]
        \includegraphics[width=\textwidth]{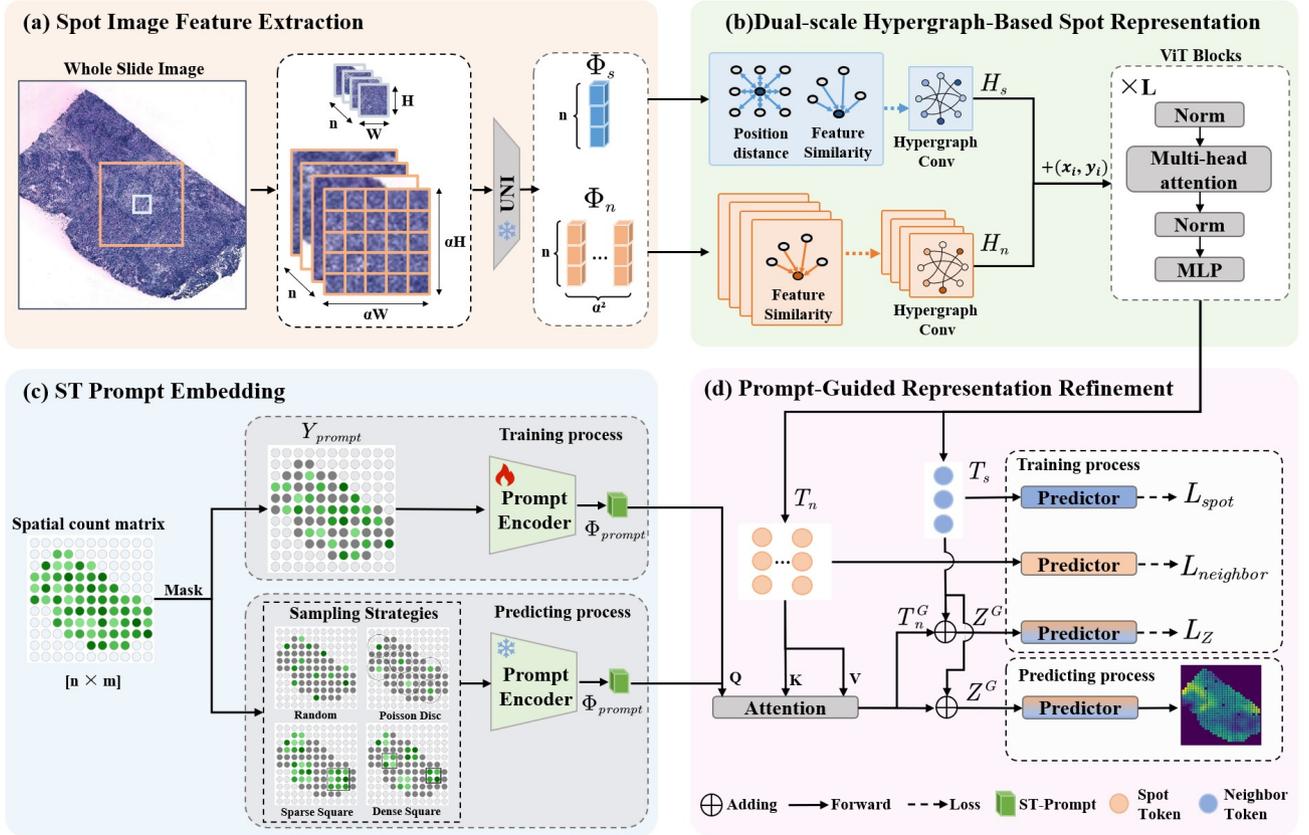}
        \caption{Overview of PH2ST.
        (a) UNI extracts histological features at two spatial scales.
        (b) Dual-scale hypergraphs capture high-order spatial relationships, with global fusion enabled by Transformer blocks.
        (c) A learnable projection module encodes the spatial gene expression matrix into ST-prompt embeddings. Four prompt sampling strategies used during inference are illustrated.
        (d) A cross-attention mechanism fuses the ST-prompt with neighboring features. The combined representations are used for final prediction.}
    \label{overview}
\end{figure*}

We formulate ST prediction as a prompt-guided multi-output regression task. Given a set of spot-level histology images from a WSI, denoted as \( X \in \mathbb{R}^{n \times H \times W \times 3} \), the objective is to predict the gene expression matrix for the unobserved spots \( Y_{\text{pred}} \in \mathbb{R}^{(n - n_{\text{prompt}}) \times m} \), guided by a small subset of spots with known expression values \( Y_{\text{prompt}} \in \mathbb{R}^{n_{\text{prompt}} \times m} \). Here, \( n \) is the total number of spots in the WSI, \( n_{\text{prompt}} \) is the number of spots with known expression, \( m \) is the number of target genes, and \( H, W \) denote the height and width of each spot image. An overview of PH2ST is shown in Fig.~\ref{overview}.
% Given a set of spot images from a WSI, denoted as $X\in \mathbb{R}^{n\times H\times W\times3} $, the goal is to predict the gene expression for each spot, represented as $ Y\in \mathbb{R}^{n\times m} $, where, $n$ is the number of spots in the WSI, $m$ is the number of genes for which are to be predicted, and $H$ and $W$ are the height and width of the spot images, respectively. 
% An overview of the proposed PH2ST is illustrated in Fig ~\ref{overview}.

% \subsection{Histology and Prompt Feature Embedding}	
\subsection{Spot Image Feature Extraction}
% The proposed PH2ST extracts both spot and neighboring histological features from each individual WSI spot image $X_i\in \mathbb{R}^{H\times W\times3}$. 
% We employ UNI \citep{chen2024uni}, a foundation pathology model, as the feature extractor, spot features $\Phi_s\in \mathbb{R}^{n\times d}$ are directly obtained from the spot image, $d$ denotes the embedding dimension.
% Neighboring features are derived from the surrounding region of the target spot. Specifically, we concatenate the features of all spots within a $5H\times 5W$ region around each target spot to obtain the neighboring features $\Phi_n\in \mathbb{R}^{n\times 25\times d}$.
PH2ST extracts both spot-level and neighborhood-level histological features from each individual WSI spot image \( X_i \in \mathbb{R}^{H \times W \times 3} \). 
We adopt UNI \citep{chen2024uni}, a universal histology image encoder pretrained on a large-scale corpus of over 100,000 WSIs spanning 20 major tissue types, as our feature extractor.
Spot-level features \( \Phi_{s} \in \mathbb{R}^{n \times d} \) are directly obtained from each spot image, where \( d \) denotes the embedding dimension.  
To capture local spatial context, neighboring features are derived from the surrounding region of each target spot. Specifically, we extract features from a \( \alpha H \times \alpha W \) region centered at each spot and concatenate the features of the $\alpha^{2}$ surrounding sub-patches to form neighborhood features \( \Phi_{n} \in \mathbb{R}^{n \times \alpha^2 \times d} \).
Note that the feature extractor can be readily adapted to other commonly used pathology-specific encoders, such as ImageNet-pretrained ResNet-50\citep{2016Deep}, CTransPath \citep{WANG2022102559}, CONCH \citep{lu2024avisionlanguage}, and Virchow \citep{zimmermann2024virchow2}.
We evaluate the effect of different encoders in the ablation section (see Section \ref{ablation}).

\subsection{ST Prompt Embedding}

To simulate limited spatial transcriptomics availability during training, we randomly mask a proportion of spots in the spatial gene expression count matrix. 
Specifically, the gene expression values of a randomly selected subset of spots are set to zero.  
We then employ a learnable prompt projection module to transform the masked expression matrix \( Y_{\text{prompt}} \) into prompt embeddings \( \Phi_{\text{prompt}} \):

\begin{equation}
    \Phi_{\text{prompt}} = \text{LN}(\text{Dropout}(\text{FC}(\beta(Y_{\text{prompt}} \cdot W)))).
\end{equation}
where \( W \) is the learnable weight matrix of the linear projection layer, \( \beta(\cdot) \) denotes the GELU activation function \citep{hendrycks2016gaussian}, \( \text{Dropout}(\cdot) \) represents the dropout operation, and \( \text{LN}(\cdot) \) denotes layer normalization.

To further simulate various real-world ST prediction usage scenarios, we adopt three prompt sampling strategies during inference: (1) fully random selection, (2) sparse or dense local square-window prompts, and (3) evenly spaced Poisson disk sampling.
These settings respectively mimic the real clinical applications, such as spatial transcriptomics imputation, local-to-global prediction, and ST super-resolution.

% We randomly mask a portion of the spots in the spatial gene expression count matrix, at the model training. Specifically, a randomly selected proportion of spots gene expression values are set to zero. We then utilize a novel defined learnable Prompt Projection module to transform this masked expression matrix $Y_{masked}$ into prompt features $\Phi_p$:
%     \begin{equation}
%         \Phi_p= LN(Dropout(FC(\beta(Y_{masked}\cdot W)))).
%     \end{equation}
% where $W$ is the linear projection operation parameter matrix, and $\beta(\cdot)$ denotes the activation function, \textcolor{red}{\textit{e.g.}}, GELU \citep{hendrycks2016gaussian}. $Dropout(\cdot)$ denotes the dropout operation and $LN(\cdot)$ denotes the layer normalization operation.

% \subsubsection{Spot and Neighboring Hypergraph Construction} \citep{bai2021hypergraph}
\subsection{Dual-scale Hypergraph-Based Spot Representation}
    % To flexibly capture complex spatial relationships among histological spots, we design a dual-scale hypergraph for spot representation learning, which models high-order dependencies among spot features at both the spot-level and neighborhood-level scales.
    We designed a Dual-scale Hypergraph-Based Spot Representation Module to model both local and global high-order relationships among histological spots via dual-scale hypergraph construction and aggregation.
    First, we construct a spot-level hypergraph among the whole slide. To simulate connections between spots with similar semantic content, we calculate the pairwise Euclidean distances between all spot features as a measure of feature similarity. The similarity between spot $v_i$ and $v_j$ is defined as: 
    \begin{equation}
        Sim(v_i,v_j)=\sqrt{\sum^d_{k=1}{(\phi_{ik}-\phi_{jk})}^2}.
    \end{equation} $\phi_{ik}$ and $\phi_{jk}$ denote the feature of $v_i$ and $v_j$ at $k_{th}$ dimension.
    
    Given that spatially adjacent spots generally exhibit similar genetic phenotypes, we establish positional distances based on the spatial coordinates of the spots. The positional distances between spot $v_i$ and $v_j$ are defined as: 
    \begin{equation}
        Pos(v_i,v_j)=\sqrt{{(x_i-x_j)}^2+{(y_i-y_j)}^2}.
    \end{equation} $(x_i,y_i)$ and $(x_j,y_j)$ denote the normalized coordinates of spot $v_i$ and $v_j$.
    
    We normalize and combine the feature similarity and positional distance to construct the incidence matrix of the hypergraph:
    \begin{equation}
    I(v_i,v_j)=Norm(Sim(v_i,v_j))+Norm(Pos(v_i,v_j)).
    \end{equation}
    where $Norm(\cdot)$ is the min-max normalization operation. Based on this incidence matrix, we identify the P nearest neighbors for each spot to construct the hyperedges.
    
    Distinct from the spot-level hypergraph construction, the neighboring-level hypergraph is constructed within the neighborhood of each spot. We directly compute the Euclidean distance between neighboring features as the incidence matrix and select the P nearest neighbors to form hyperedges. It is important to note that the neighboring-level hypergraph comprises $n$ sub-hypergraphs. We then employ HypergraphConv \citep{bai2021hypergraph} to aggregate spot features with their shared hyperedge spots, using the values of the incidence matrix as hyperedge weights. The resulting spot and neighboring features after hypergraph processing can be represented as:
    \begin{equation}
        HypergraphConv(\Phi,I)=D^{-1}IW_HD^{-1}I^T\Phi,
    \end{equation}
    \begin{equation}H_s=Dropout(Norm(\sigma(HypergraphConv(\Phi_s,I))).
    \end{equation}
    \begin{equation}
        H_n=Dropout(Norm(\sigma(HypergraphConv(\Phi_n,I))).
    \end{equation}
    where $\sigma(\cdot)$ denotes the ReLU activation function \citep{agarap2018deep}. $W_H$ is the diagonal hyperedge weight matrix, and $D$ is the corresponding degree matrix.

    While graph neural networks excel at fusing neighboring spot features, they are limited in capturing long-range dependencies between spots. To further integrate the global representation of spot features, we apply Vision Transformer (ViT) blocks to the spot and neighboring features after hypergraph convolution. 
    % Multi-head attention is a linear combination of the outputs of multiple attention heads. 
    The multi-head attention mechanism allows our model to effectively model feature interactions between spot and neighboring representations at the global level.
    The resulting representations of the spot and neighboring tokens are $T_s \in \mathbb{R}^{n\times d}$ and $T_n \in \mathbb{R}^{n\times d}$. Within the Vision Transformer architecture, the attention mechanism for each spot image is learned via a multi-head self-attention mechanism, resulting in distinct weights being assigned to all patches:
    \begin{equation}
        Attention(Q,K,V)=softmax(\frac{QK^T}{\sqrt{d_k}})V.
    \end{equation}
    \begin{equation}
        head_i=Attention(H_sW^Q_i,H_sW^K_i,H_sW^V_i)
    \end{equation}
    \begin{equation}
        T_{s/n}=concat(head_1,head_2,...,head_h)\times W^h
    \end{equation}
    where $Q,K,V$ denotes Query, Key and Value, respectively; $W^Q_i, W^K_i, W^V_i$ are the weight matrices of $Q,K,V$ of $i_{th}$ head; $\sqrt{d_k}$ is used to controls the scaling of the attention weights by adjusting the denominator in the dot-product attention mechanism. $W^h$ is the weight matrix of multiple attention heads and $h$ is the number of attention heads.
    
    % The spot representation, derived from the slide-level hypergraph, incorporates more flexible and specific semantic information. While, the neighbor-level hypergraph, operating within the spot neighborhood, yields a smoother representation. 
    The spot-level representation, derived from the slide-level hypergraph, captures long-range dependencies while preserving the unique semantic characteristics of each spot.
    In contrast, the neighbor-level hypergraph, constrained to local regions, yields smoother and more stable features.
    These two representations complement each other, enhancing the model’s ability to capture fine-grained details while preserving local consistency.

% \subsection{Integrating ST-Prompts and Histological Hypergraphs}
\subsection{Prompt-Guided Representation Refinement and Gene Expression Prediction}
To fully exploit the informative signals embedded in sparse ST-prompt data, we introduce a cross-attention mechanism to guide the refinement of spot representations for improved gene expression prediction. Specifically, features derived from the ST-prompt serve as queries, while the neighboring histological features extracted from the WSI act as keys and values in the attention computation. 
The resulting attention-guided features are then integrated via an additive operation to obtain the final fused representation:
    % To better capture the relationship between histological visual features and spatial gene expression by leveraging ST-prompt guidance, we implement a cross-attention layer. The features derived from the ST-prompt serve as query, while the neighboring features from the WSI serve as keys and values for the attention mechanism. The attention-guided features are then integrated through an additive operation to obtain the final feature representation:
    \begin{equation}
        T_n^G = CrossAttn(\Phi_{prompt},T_n).
    \end{equation}
    \begin{equation}
        Z^G=Sum(T_n^G,T_s).
    \end{equation}
    where $T_n^G$ indicates the guided neighboring features, and $Z^G$ denotes the integrated feature representation. By means of this mechanism, the ST-prompt facilitates the iterative and localized refinement of neighboring features, culminating in a global update across the entire image.
    
    We calculate the MSE loss between fused prediction and the labels, then use a weighted sum to obtain the final loss:
    \begin{equation}
        \mathcal{L}_Z=\frac{1}{m}\sum^{m}_c{(p^c_Z-g^c)}^2.
    \end{equation}
    where $g^c$ and $p_Z^c$ represent the label and prediction of the $c_{th}$ gene ($c\in \{1,...,m\}$) obtained by a fully connected layer to $Z^G$. We calculated the Mean Squared Error (MSE) loss between the predictions derived from the spot and its neighbors and the sum of the labels. We also calculated the MSE loss between these predictions and the fused prediction result. The final loss is balanced the different losses with a hyperparameter $\lambda$:
    \begin{align*}
        \mathcal{L}_{spot}=(1-\lambda)\frac{1}{m}\sum^{m}_c{(p^c_{spot}-g^c)}^2+ \\
        \lambda\frac{1}{m}\sum^{m}_c{(p^c_{spot}-p^c_Z)}^2,
    \end{align*}
    \begin{align*}
        \mathcal{L}_{neighbor}=(1-\lambda)\frac{1}{m}\sum^{m}_c{(p^c_{neighbor}-g^c)}^2+ \\
         \lambda\frac{1}{m}\sum^{m}_c{(p^c_{neighbor}-p^c_Z)}^2,
    \end{align*}
    \begin{equation}
        \mathcal{L}_{final}= \mathcal{L}_{spot}+\mathcal{L}_{neighbor}+\mathcal{L}_Z.
    \end{equation}
    where $p^c_{spot}$ and $p^c_{neighbor}$ represent the spot and neighboring prediction obtained by a fully connected layer to $T^G_n$ and $T_l$ respectively.

\section{Experiments and Results}
    \subsection{Experimental Settings}
    \subsubsection{Dataset}
    We utilized two widely used spatial transcriptomics datasets from the 10x Genomics platform: human HER2-positive breast cancer (HER2+) \citep{andersson2021spatial} and human cutaneous squamous cell carcinoma (cSCC) \citep{ji2020multimodal}. The HER2+ dataset comprises 36 samples from 8 breast cancer patients, while the cSCC dataset includes 12 samples from 4 patients. Each spot within the selected dataset is subjected to a cropping operation, resulting in a 224$\times$224 pixel patch. We considered only the top 1,000 highly variable genes for each dataset \citep{pang2021leveraging}. Gene counts were then normalized by dividing by the total count across all genes, multiplying by 1,000,000, and applying a natural log transformation. After processing, the HER2+ dataset contained 785 genes, and the cSCC dataset contained 171 genes. The total counts of spots in HER2+ and cSCC datasets are 13,620 and 23,205.

    \subsubsection{Implementation Details}
    % \noindent\textbf{Implementation details.}
    % For WSI preprocessing, each spot was cropped into a 224×224 pixel patch.
    % Neighboring spots were extracted by cropping a 1120×1120 image centered on the current spot and dividing it into 25 equal-sized patches.
    For WSI preprocessing, each spot corresponds to a 224×224 pixel image patch. $\alpha$ was set to 5. To capture spatial context, a 1120×1120 region centered on each spot was extracted and uniformly divided into 25 non-overlapping 224×224 patches, representing the spot’s neighborhood.
    Experiments were conducted with 5-fold cross-validation.
    For PH2ST, the hyperparameter $\lambda$ was set to 0.3, and $P$ was set to 4. 
    The model was optimized using the Adam optimizer \citep{kingma2014adam} with an initial learning rate 0.0001 and dynamic adjustment strategy (step size = 50, decay rate = 0.9). 
    The number of training epochs was set to 200, with an early stopping mechanism triggered if there was no improvement in PCC after 20 epochs. 
    During training, PH2ST randomly selected 30\% of the spots with ST labels per WSI as prompts in each iteration. 
    During testing, only 10\% of the spots are randomly selected for prompt generation.
    For the baseline methods, we follow the training settings from their original papers but additionally use the same amount of 10\% spots with ST labels of each test case to fine-tune them before evaluation to ensure a fair comparison. 
    All experiments were implemented in PyTorch and PyTorch-Lightning and trained on an NVIDIA RTX 4090 GPU.

    \begin{figure*}[t]
    \centering
    \includegraphics[width=\textwidth]{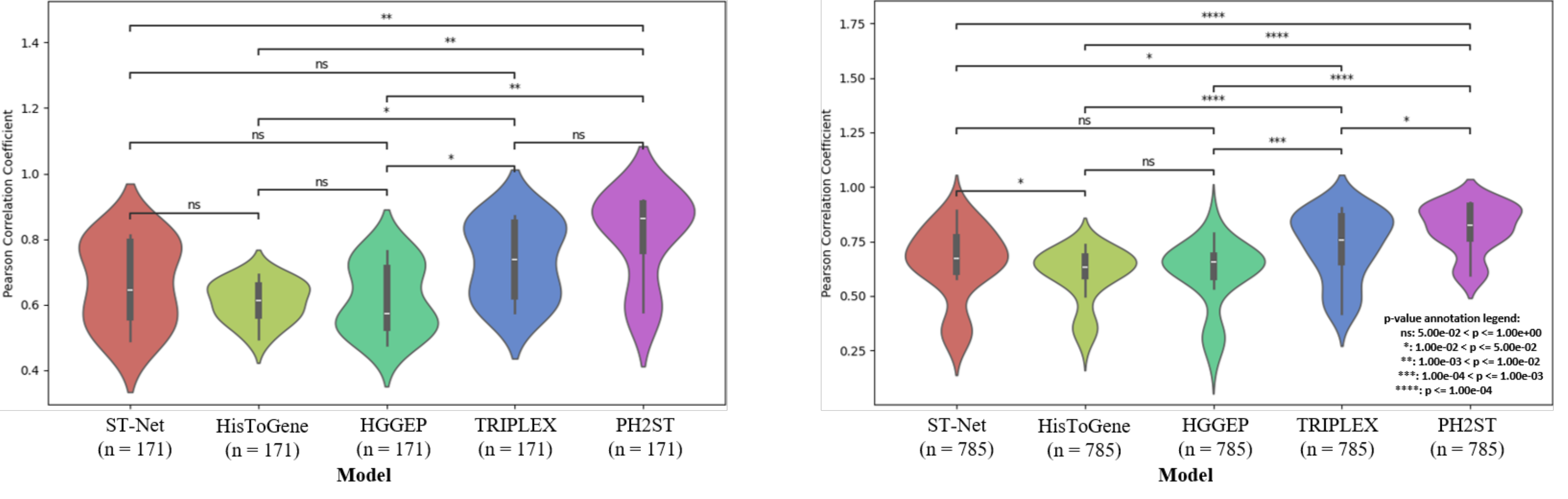}
    \caption{\centering {Comparison of PCC on HER2+ and cSCC datasets using non-parametric test.}}
    % Prediction visualizations on ST dataset. The visualizations include ground truth and predicted gene expression levels for cSCC-related genes from ST-Net, HisToGene, HGGEP, TRIPLEX, and PH2ST.
    % } 
    \label{violin}
    \end{figure*}

    \begin{table*}
    \renewcommand{\arraystretch}{1.2}
    \footnotesize
    \caption{\centering Performance of different methods on HER2+ and cSCC ST datasets. The best results are highlighted in bold.}
    \label{tab1}
    \centering
    \begin{tabular}{c *{6}{wc{1.8cm}}}  % 增加列宽
        \toprule
    % ...  表格的其他部分 (使用 \midrule 和 \bottomrule)
         \multirow{2}{*}{Model} & \multicolumn{3}{c}{HER2+}                                                              & \multicolumn{3}{c}{cSCC}                                                               \\ \cmidrule(lr){2-4} \cmidrule(lr){5-7} 
                       & \multicolumn{1}{c}{MAE (↓)} & \multicolumn{1}{c}{CCC (↑)} & \multicolumn{1}{c}{PCC (↑)} & \multicolumn{1}{c}{MAE (↓)} & \multicolumn{1}{c}{CCC (↑)} & \multicolumn{1}{c}{PCC (↑)} \\ \midrule
    ST-Net \citep{he2020integrating}                 & \multicolumn{1}{c}{0.654$\pm$0.065}       & \multicolumn{1}{c}{0.129$\pm$0.092}       & \multicolumn{1}{c}{0.187$\pm$0.116}       & \multicolumn{1}{c}{0.699$\pm$0.125}       & \multicolumn{1}{c}{0.248$\pm$0.048}       & \multicolumn{1}{c}{0.307$\pm$0.052}        \\
    HisToGene \citep{pang2021leveraging}              & \multicolumn{1}{c}{0.656$\pm$0.054}       & \multicolumn{1}{c}{0.104$\pm$0.058}       & \multicolumn{1}{c}{0.176$\pm$0.092}       & \multicolumn{1}{c}{1.081$\pm$0.105}       & \multicolumn{1}{c}{0.079$\pm$0.029}       & \multicolumn{1}{c}{0.238$\pm$0.047}       \\
    Hist2ST \citep{zeng2022spatial}                & \multicolumn{1}{c}{0.679$\pm$0.075}       & \multicolumn{1}{c}{0.106$\pm$0.070}       & \multicolumn{1}{c}{0.201$\pm$0.117}                            & \multicolumn{1}{c}{1.076$\pm$0.108}       & \multicolumn{1}{c}{0.078$\pm$0.044}       & \multicolumn{1}{c}{0.224$\pm$0.067}                            \\
    HGGEP \citep{li2024gene}                  & \multicolumn{1}{c}{0.643$\pm$0.067}       & \multicolumn{1}{c}{0.123$\pm$0.081}       & \multicolumn{1}{c}{0.190$\pm$0.094}                            & \multicolumn{1}{c}{0.736$\pm$0.161}       & \multicolumn{1}{c}{0.125$\pm$0.096}       & \multicolumn{1}{c}{0.173$\pm$0.084}                            \\
    BLEEP \citep{xie2024spatially}                  & \multicolumn{1}{c}{0.742$\pm$0.049}       & \multicolumn{1}{c}{0.177$\pm$0.010}       & \multicolumn{1}{c}{0.202$\pm$0.046}                            & \multicolumn{1}{c}{0.913$\pm$0.101}       & \multicolumn{1}{c}{0.170$\pm$0.009}       & \multicolumn{1}{c}{0.174$\pm$0.008}                            \\
    TRIPLEX \citep{chung2024accurate}                & \multicolumn{1}{c}{0.543$\pm$0.110}       & \multicolumn{1}{c}{0.225$\pm$0.131}       & \multicolumn{1}{c}{0.318$\pm$0.147}                            & \multicolumn{1}{c}{0.702$\pm$0.306}       & \multicolumn{1}{c}{0.263$\pm$0.130}       & \multicolumn{1}{c}{0.431$\pm$0.060}                            \\
\rowcolor{light-gray}    PH2ST                 & \multicolumn{1}{c}{\textbf{0.516$\pm$0.096}}       & \multicolumn{1}{c}{\textbf{0.256$\pm$0.148}}       & \multicolumn{1}{c}{\textbf{0.338$\pm$0.139}}                            & \multicolumn{1}{c}{\textbf{0.698$\pm$0.320}}       & \multicolumn{1}{c}{\textbf{0.312$\pm$0.147}}       & \multicolumn{1}{c}{\textbf{0.479$\pm$0.103}}                            \\ \bottomrule
    \end{tabular}
    \end{table*}
    
    \subsubsection{Evaluation Metrics}
    % Evaluation metrics include Mean Absolute Error (MAE), Concordance Correlation Coefficient (CCC), and Pearson Correlation Coefficient (PCC).
    To assess the degree of correlation between the predictions and the ground truth, we use the Pearson Correlation Coefficient (PCC). The PCC, which ranges from -1 (perfect negative correlation) to 1 (perfect positive correlation), is computed by dividing the covariance of the two variables by the product of their standard deviations:
    \begin{equation}
        PCC=\frac{Cov(X_{true},X_{pred})}{\sigma(X_{true})\cdot\sigma(X_{pred})}.
    \end{equation}
    where $Cov(\cdot)$ denotes covariance; $X_{true}$ and $X_{pred}$ denote the ground truth and the prediction, respectively. $\sigma(\cdot)$ denotes standard deviation.
    
    To assess the level of agreement between the two variables, we employ the Concordance Correlation Coefficient (CCC). The CCC, which takes values between -1 and 1, provides a measure of how well the clustering results align with the true labels, with values closer to 1 signifying stronger agreement. The CCC is calculated as follows:
    \begin{equation}
        CCC=\frac{2\rho\sigma(X_{true})\sigma(X_{pred})}{{\sigma(X_{true})}^2+{\sigma(X_{pred})}^2+{(\mu_{X_{true}}-\mu_{X_{pred}})}^2}.
    \end{equation}
    where $\rho$ denotes the correlation coefficient; $\mu_{X_{true}}$ and $\mu_{X_{pred}}$ denote the mean value of the ground truth and the prediction, respectively.

    The Mean Absolute Error (MAE) is defined as the average of the absolute differences between predicted and observed values. The MAE can be computed as follows:
    \begin{equation}
        MAE = \frac{1}{m}\sum^m_{i=1}|X_{pred}-X_{true}|.
    \end{equation}
    where $m$ denotes the number of predicted genes.

    \begin{figure*}[ht]
    \centering
    \includegraphics[width=0.9\textwidth]{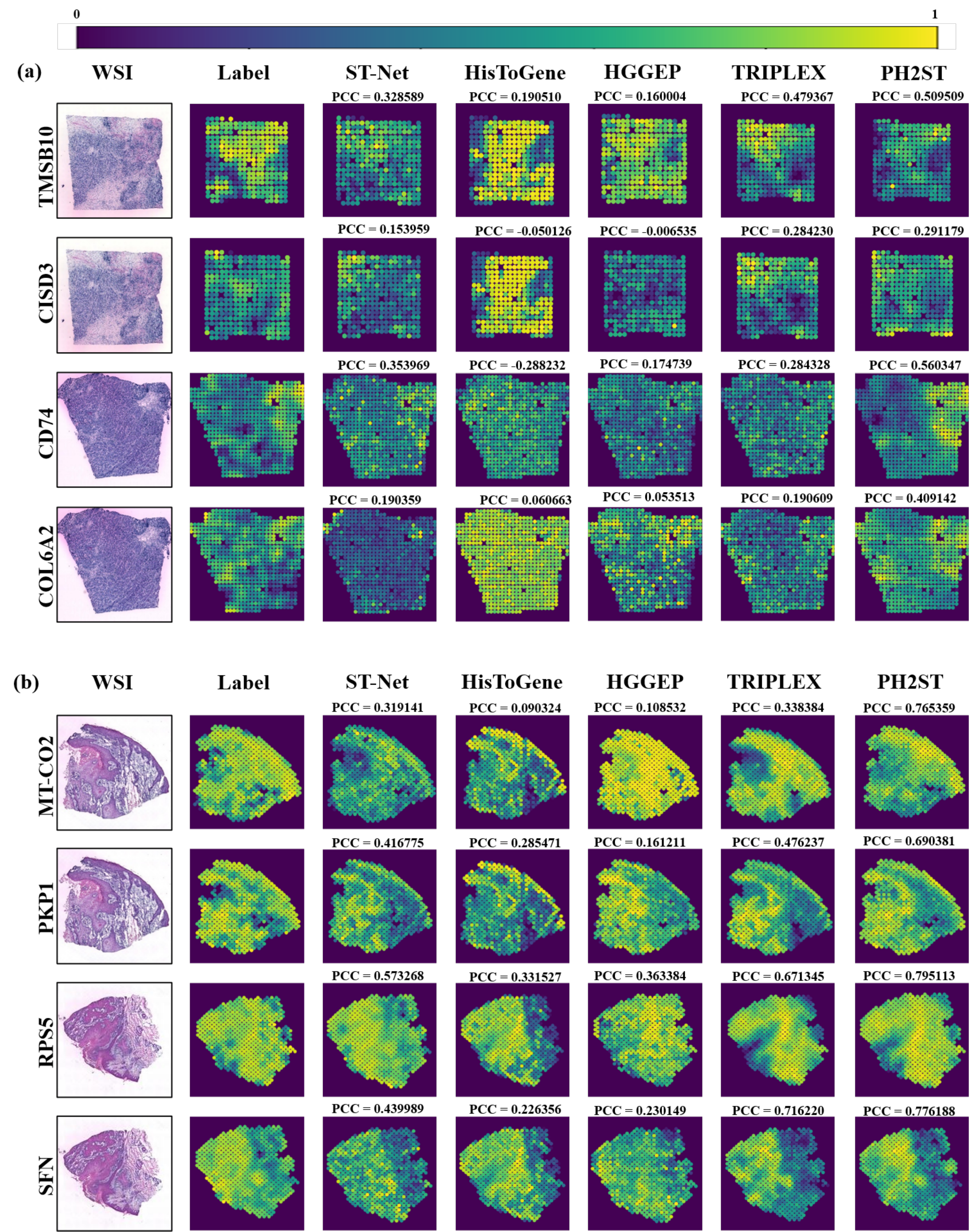}
    \caption{
    Visualization of WSI thumbnails, ST ground truths, and model predictions on two datasets. A representative sample was randomly selected from each dataset.
    (a) Marker genes: $\textit{TMSB10}$, $\textit{CISD3}$, $\textit{CD74}$ and $\textit{COL6A2}$ on HER2+ dataset. 
    (b) Marker genes: $\textit{MT-CO2}$, $\textit{PKP1}$, $\textit{RPS5}$ and $\textit{SFN}$ on cSCC dataset.
    % Prediction visualizations on ST dataset. The visualizations include ground truth and predicted gene expression levels for cSCC-related genes from ST-Net, HisToGene, HGGEP, TRIPLEX, and PH2ST.
    } \label{visual}
    \end{figure*}

    \begin{figure*}[t]
    \centering
    \includegraphics[width=\textwidth]{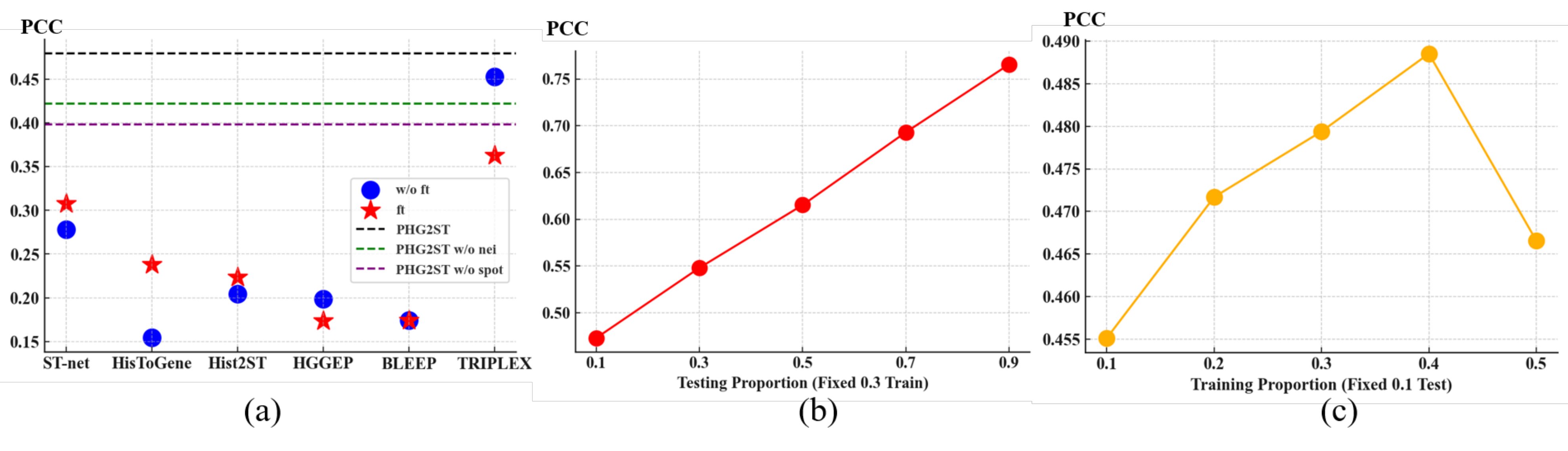}
    \caption{Ablation results (PCC and CCC) on the cSCC dataset.
    (a) Effect of fine-tuning on baseline methods and contribution of the proposed dual-scale histological hypergraphs.
    (b–c) Influence of ST-prompt proportions during training and testing on model performance.} \label{ablation}
    \end{figure*}
    
    \subsection{Experimental Results and Visualization}
    % \subsubsection{Results of Benchmark Comparison}
    We compared PH2ST to six representative methods on the HER2+ and cSCC datasets in Table~\ref{tab1}. 
    ST-Net \citep{he2020integrating} and HisToGene \citep{pang2021leveraging} are spot-based methods, BLEEP \citep{xie2024spatially} utilizes contrastive learning to co-embed spot images and gene expression, Hist2ST \citep{zeng2022spatial} and HGGEP \citep{li2024gene} utilize graph neural networks and hypergraph networks, and TRIPLEX \citep{chung2024accurate}, the current state-of-the-art, adopts a multi-scale feature fusion strategy for enhanced prediction performance.
    PH2ST achieved the best performance across all evaluation metrics compared to all baseline methods. Except for TRIPLEX, it outperformed the remaining methods by at least 7\% in CCC and 14\% in PCC on both datasets. Even when compared with TRIPLEX, the current state-of-the-art, PH2ST still demonstrated superior performance, with improvements of 3–5\% in CCC and 2–5\% in PCC across the two datasets.
    These results demonstrate the effectiveness of PH2ST in ST prediction under this inference-time prompting setting.

    To assess statistical significance, we conducted a non-parametric test on PCC scores from five competing models across the two datasets. 
    As shown in Fig.~\ref{violin}, PH2ST demonstrates statistically superior performance compared to all baselines.
    Furthermore, qualitative visualizations (Fig.~\ref{visual}) revealed that PH2ST predictions closely align with the ground truth in terms of expression distribution patterns.
    
    % \begin{figure*}[t]
    % \centering
    % \includegraphics[width=\textwidth]{sampling.pdf}
    % \caption{Visual representation of four sampling strategies.
    % % Prediction visualizations on ST dataset. The visualizations include ground truth and predicted gene expression levels for cSCC-related genes from ST-Net, HisToGene, HGGEP, TRIPLEX, and PH2ST.
    % } 
    % \label{sampling_fig}
    % \end{figure*}

    \subsection{Ablation}
    We conducted ablation studies to evaluate (1) the impact of fine-tuning for baseline models, (2) the effectiveness of multi-scale histological hypergraph features, (3) the influence of ST-prompt proportions during training and testing, and (4) the effectiveness of using different feature extractors for spot images on the prediction performance.
    
    % since we use a small subset of ST labels as prompts to guide the prediction of spatially resolved gene expression, To ensure a fair comparison with other methods, 
    First, we freeze all layers of the baseline models except the final linear layer and fine-tune them for five epochs using the same small subset of ST labels from the test set. BLEEP is excluded from fine-tuning due to its contrastive learning architecture.
    As shown in Fig.~\ref{ablation} (a), fine-tuning improves the performance of some baseline models, particularly spot-based methods, by adapting them to test ST data. However, for others, architectural limitations render fine-tuning on sparse spots ineffective or even harmful, resulting in degraded performance.
    
    Secondly, Fig.~\ref{ablation} (a) (dashed lines) shows that integrating multi-scale histological hypergraph features, from cellular morphology to surrounding tissue phenotypes, improves spatial gene expression prediction.
    Using either the spot branch or the neighboring branch in isolation leads to a decrease in model performance.
    
    Moreover, Fig.~\ref{ablation} (b) illustrates that higher ST-prompt ratios at inference improve performance, confirming the framework’s ability to integrate prompt information.
    However, Fig.~\ref{ablation} (c) shows that excessive ST-prompt use during training results in diminishing returns, likely due to over-reliance on prompts over histological features.
    
    Lastly, we evaluated five different spot image feature extractors in this ablation. 
    CTransPath \citep{WANG2022102559} is a Transformer-based unsupervised model; Virchow2 \citep{zimmermann2024virchow2} is a self-supervised Vision Transformer pretrained on whole slide images; and UNI \citep{chen2024uni} is a general-purpose vision encoder trained specifically for histopathology. CONCH1.5 \citep{lu2024avisionlanguage} is a vision-language foundation model designed for computational pathology. 
    The experimental results, summarized in Table~\ref{table3}, indicate that feature extractors pretrained on large-scale histopathology datasets consistently achieve better performance in ST prediction, with UNI achieving the best overall results.

    \begin{table}[]
    \centering
    \caption{Performance comparison using different spot encoder on HER2+ dataset. The best results are highlighted in bold.}
    \label{table3}
    \begin{tabular}{cccc}
        \hline
        Encoder    & MAE (↓)                   & CCC (↑)                   & PCC (↑)                   \\ \hline
        ResNet50   & 0.647$\pm$0.089          & 0.077$\pm$0.014          & 0.180$\pm$0.023          \\
        Ctranspath & 0.536$\pm$0.107          & 0.198$\pm$0.129          & 0.320$\pm$0.101          \\
        CONCH1.5   & 0.526$\pm$0.097          & 0.201$\pm$0.100          & 0.325$\pm$0.126          \\
        Virchow2   & 0.560$\pm$0.054          & 0.198$\pm$0.101          & 0.338$\pm$0.129          \\
\rowcolor{light-gray}        UNI        & \textbf{0.516$\pm$0.096} & \textbf{0.256$\pm$0.148} & \textbf{0.338$\pm$0.139} \\ \hline
    \end{tabular}
    \end{table}

    \begin{table*}[]
    \centering
    \caption{\centering Performance comparison using different sampling method on HER2+ datasets. The best results are highlighted in bold.}
    \label{tab2}
    \begin{tabular}{cccccc}
    \hline
    \multirow{2}{*}{MAE (↓)} & \multicolumn{5}{c}{Select Ratio}                                                                                                     \\
                            & 0.1                      & 0.3                      & 0.5                      & 0.7                      & 0.9                      \\ \hline
    Random                  & \textbf{0.516$\pm$0.096} & \textbf{0.487$\pm$0.090} & \textbf{0.455$\pm$0.084} & 0.423$\pm$0.079          & \textbf{0.391$\pm$0.076} \\
    Poisson Disc Sampling   & 0.520$\pm$0.097          & 0.494$\pm$0.091          & 0.471$\pm$0.086          & 0.454$\pm$0.084          & 0.439$\pm$0.082          \\
    Sparse Sqaure Sampling  & 0.521$\pm$0.097          & 0.488$\pm$0.090          & 0.457$\pm$0.084          & 0.424$\pm$0.078          & 0.393$\pm$0.075          \\
    Dense Square Sampling   & 0.519$\pm$0.097          & 0.487$\pm$0.090          & 0.455$\pm$0.084          & \textbf{0.422$\pm$0.079} & 0.391$\pm$0.076          \\ \hline
    \multirow{2}{*}{CCC (↑)} & \multicolumn{5}{c}{Select Ratio}                                                                                                     \\
                            & 0.1                      & 0.3                      & 0.5                      & 0.7                      & 0.9                      \\ \hline
    Random                  & 0.214$\pm$0.117          & 0.267$\pm$0.112          & \textbf{0.322$\pm$0.106} & \textbf{0.379$\pm$0.103} & 0.441$\pm$0.095          \\
    Poisson Disc Sampling   & 0.213$\pm$0.116          & 0.256$\pm$0.114          & 0.295$\pm$0.109          & 0.324$\pm$0.107          & 0.350$\pm$0.106          \\
    Sparse Sqaure Sampling  & 0.211$\pm$0.117          & 0.265$\pm$0.113          & 0.319$\pm$0.108          & 0.424$\pm$0.078          & 0.393$\pm$0.075          \\
    Dense Square Sampling   & \textbf{0.215$\pm$0.117} & \textbf{0.268$\pm$0.113} & 0.321$\pm$0.108          & 0.379$\pm$0.103          & \textbf{0.442$\pm$0.097} \\ \hline
    \multirow{2}{*}{PCC (↑)} & \multicolumn{5}{c}{Select Ratio}                                                                                                     \\
                            & 0.1                      & 0.3                      & 0.5                      & 0.7                      & 0.9                      \\ \hline
    Random                  & \textbf{0.338$\pm$0.143} & \textbf{0.398$\pm$0.121} & \textbf{0.459$\pm$0.102} & \textbf{0.526$\pm$0.086} & 0.606$\pm$0.064          \\
    Poisson Disc Sampling   & 0.336$\pm$0.143          & 0.384$\pm$0.127          & 0.429$\pm$0.111          & 0.462$\pm$0.104          & 0.492$\pm$0.097          \\
    Sparse Sqaure Sampling  & 0.334$\pm$0.144          & 0.395$\pm$0.122          & 0.455$\pm$0.105          & 0.522$\pm$0.086          & 0.601$\pm$0.066          \\
    Dense Square Sampling   & 0.338$\pm$0.143          & 0.398$\pm$0.122          & 0.459$\pm$0.104          & 0.526$\pm$0.087          & \textbf{0.607$\pm$0.067} \\ \hline
    \end{tabular}
    \end{table*}

    \subsection{Evaluation under Real-World Prompt Sampling Strategies}
    % Current ST sequencing technologies predominantly utilize square-shaped chips of varying sizes \citep{Moffitt2022the}. Selecting an appropriate chip size is crucial for gaining a comprehensive understanding of cellular and tissue-level changes during carcinogenesis. This work suggests that the inclusion of even a small subset of ST data can improve the accuracy of spatial transcriptomics predictions. 
    Current ST platforms, such as 10x Visium, are limited by relatively high dropout rates and spatial resolutions that fall short of single-cell granularity. Moreover, they typically rely on square-shaped chips of varying sizes \citep{Moffitt2022the}, which in practice are used to profile only one or a few local tissue regions per WSI.
    These practical constraints highlight the need for effective prompt selection strategies that align with real-world usage patterns. To this end, we designed four representative prompt sampling strategies that simulate common application scenarios:
    (1) Fully random selection, to mimic unpredictable dropout or irregular sparse measurements;
    (2) Poisson disc sampling to simulate low-resolution ST measurements with even spatial coverage, corresponding to super-resolution prediction tasks;
    (3) Sparse local square-window sampling, modeling the use of a large-area chip covering part of the tissue at low density; and
    (4) Dense local square-window sampling, representing a small-area chip covering part of the tissue at high density.
    % These sampling strategies allow us to evaluate PH2ST’s robustness and generalization across realistic application scenarios. 
    % To further explore this, we conducted experiments using various prompt selection methods to determine which prompts are most effective in capturing relevant tissue phenotypes and gene associations. 
    % Four different spot sampling strategies were employed: random sampling, Poisson disc sampling, uniform sampling within a large square region (Sparse Square Sampling), and uniform sampling within a small square region (Dense square sampling). Poisson disc sampling provides a spatially more uniform distribution of points while maintaining randomness. The square area sampling strategies mimic the sequencing process of spatial transcriptomics chips of varying sizes. 
    % Figure \ref{sampling_fig} provides a visual representation of these four sampling strategies. 
    % The experimental results 
    
    The results are summarized in Table \ref{tab2}, showing a clear trend: as the proportion of prompt spots increases, the model’s prediction performance improves consistently.
    More importantly, compared to random sampling, all three other prompt sampling strategies—Poisson disc, sparse square, and dense square—demonstrate competitive performance in most settings, confirming PH2ST’s robustness and generalizability across various practical applications.
    However, notable differences emerge, especially under higher ST-prompt proportions. Poisson disc sampling, which simulates evenly spaced low-resolution measurements, yields the weakest performance. We hypothesize this is due to an ``island effect'', where isolated prompts lack neighboring context, limiting the prompt encoder’s ability to generate semantically meaningful representations.
    In contrast, dense square sampling (small-area, high-density) outperforms sparse square sampling (large-area, low-density), suggesting that denser local regions provide more informative and cohesive context for downstream prediction. Sparse sampling may inadvertently include low-informative regions (\textit{e.g.}, adipose or stroma), reducing prompt effectiveness.
    These findings suggest that for a fixed ST measurement area, using smaller chips in multiple informative regions may be more beneficial than covering a large but sparse region. Moreover, the relatively poor performance in the super-resolution (Poisson disc) setting indicates that future work is needed to further improve PH2ST’s capability in low-resolution extrapolation tasks.
    
    % demonstrating that gradually increasing the proportion of data used for ST-prompting improves the model's predictive performance. 
    % Random sampling and dense square sampling generally yield better predictions compared to the other two sampling strategies. 
    % Sparse square sampling performs slightly worse, while Poisson disc sampling exhibits the poorest performance. 
    % We hypothesize that random and dense square sampling, while covering a sufficient area of the tissue of interest, also allows the model to incorporate gene information from neighboring spots, effectively leveraging the information fusion capabilities of the hypergraph neural network. 
    % Sparse square sampling may inadvertently select non-cancerous or low-cancer-related regions (\textcolor{red}{\textit{e.g.}}, adipose tissue), thereby diminishing the guidance provided by the prompts. The weaker performance of Poisson disc sampling is likely attributable to an ``island effect'' caused by the lack of gene expression information from neighboring tissues, hindering the prompt encoder's ability to generate semantically rich prompts.

    \begin{figure*}[t]
    \centering
    \includegraphics[width=\textwidth]{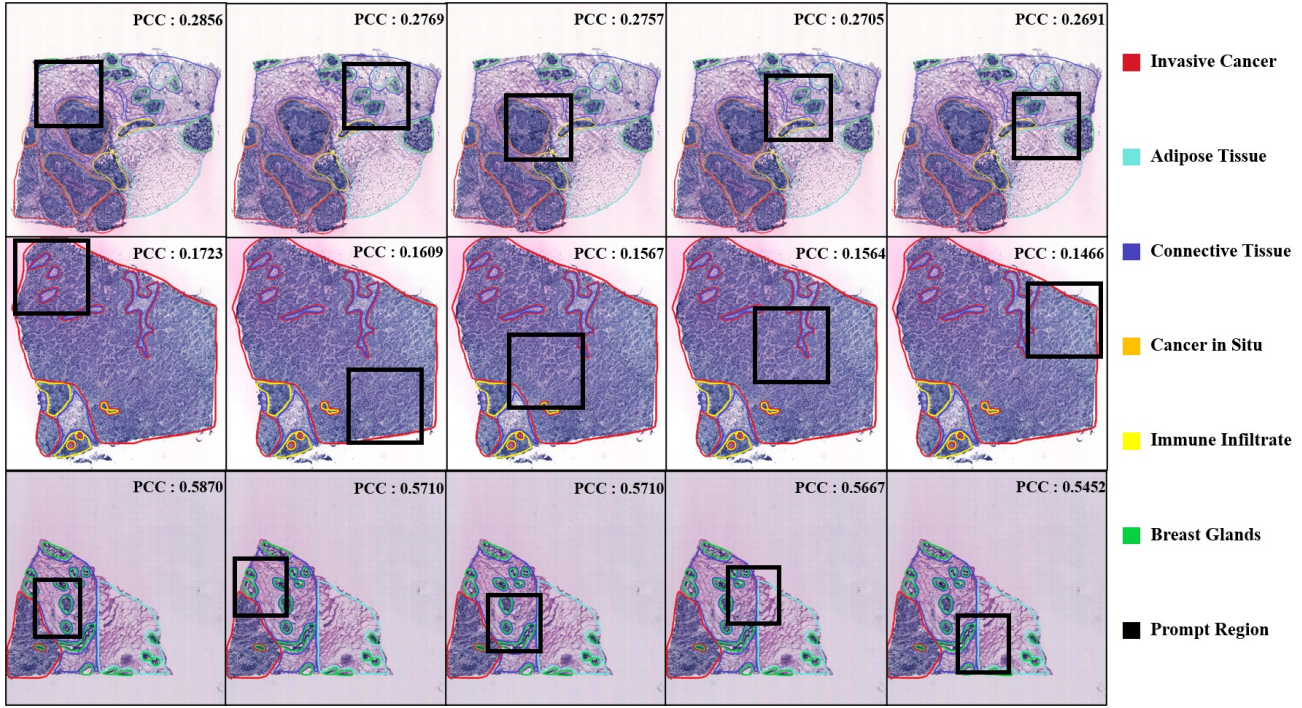}
    \caption{
    Effect of different prompt region selections on ST prediction performance (PCC). The black boxes indicate the selected prompt regions, while the colored outlines represent pathologist-annotated tissue types.
    % Visualization of expert pathologist annotations and prompt selection regions within slices H1, F1, and B1 of the HER2+ dataset.
    } \label{region}
    \end{figure*}

    \section{Discussion}\label{discussion}
    
    \subsection{Clinical applications}
    PH2ST demonstrated strong potential in addressing three key challenges that currently limit the large-scale application of ST:
    \begin{itemize}
    \item \textit{High prevalence of missing values.}
    Inefficient mRNA capture, transcriptional bursting, and RNA degradation often result in substantial dropout in ST data, which can severely hinder downstream analyses \citep{wen2023single}. Traditional imputation methods frequently lead to over-smoothed results and lack generalizability \citep{li2024tissue}. In contrast, our prompt-guided PH2ST requires only partial ST measurements to model global spatial gene expression distributions and consistently outperforms state-of-the-art methods across a wide range of missingness (10\%–90\%).
    \item \textit{Inflexibility of chip size and capture regions.}
    The effective capture area of ST chips varies across platforms, ranging from tens to hundreds of micrometers \citep{wirth2023spatial,xu2024dna}. PH2ST flexibly adapts to these variations by simulating different chip sizes through targeted prompt sampling. For example, Sparse and Dense Square Sampling strategies emulate chips with large and small capture areas, respectively—both yielding stable and consistent prediction performance and underscoring the method’s robustness across chip sizes.
    \item \textit{Low-resolution ST and super-resolution prediction.}
    Many mainstream ST platforms, such as 10x Visium, offer relatively coarse resolution that falls short of the single-cell level. PH2ST shows promising capability for ST super-resolution by accurately inferring high-resolution expression patterns from sparse, low-resolution inputs. Notably, Poisson disk-based sampling with low prompt ratios (\textit{e.g.}, 0.1 or 0.3) demonstrates comparable robustness to other sampling strategies, suggesting the framework’s potential for cost-effective, high-resolution ST reconstruction.
    \end{itemize}

    \subsection{Insights into Prompt Region Selection}
    % \textcolor{red}{We posit that tissues exhibiting a combination of relevance to the target tumor type, high data quality, and rich informative features are most suitable for prompt encoding in ST prediction. We conducted experiments using sparse square sampling to randomly sample regions from pathology images, as illustrated in Figure~\ref{region}. The results indicate that regions with high tumor cell density or distinctive microenvironment characteristics are particularly effective for prompt encoding.}
    As mentioned before, in real-world ST applications, chip size limitations typically restrict measurement to only one or a few local regions within a WSI. 
    This raises a critical question: Which tissue regions are most informative for guiding accurate ST prediction across the remaining areas?
    To explore this, we conducted a sliding-window experiment in which prompt regions of equal size were sampled from different locations within each WSI. These regions were used as ST prompts to guide PH2ST, and we evaluated the resulting prediction performance.
    % examples from three WSIs demonstrate the variability in prediction outcomes. The results show that regions with high tumor cell density or distinctive microenvironmental features consistently lead to more accurate predictions, underscoring their value as informative prompt regions.
    
    Our findings indicate a strong correlation between the choice of tissue-specific prompts and the accuracy of ST prediction, as illustrated in Fig.~\ref{region}. Specifically, prompts derived from tumor boundaries that include surrounding stromal regions or well-defined transitions between tissue types tend to yield more accurate predictions. In contrast, using prompts from regions with sparse tumor cells or from areas composed entirely of tumor tissue without microenvironmental context leads to decreased performance. These results suggest that moderate tissue heterogeneity—particularly at the tumor-stroma interface—provides richer contextual cues that facilitate the learning of genotype–phenotype associations, making such regions more effective for prompt-based prediction.

    \subsection{Limitations and Future Works}
    Despite the strong performance of our model in predicting global spatial gene expression, some limitations remain. 
    The hypergraph construction process, performed during training, cannot be GPU-accelerated \citep{feng2019hypergraph}, leading to increased computational cost. 
    Furthermore, due to current data limitations, many experimental settings in this study remain simulation-based. 
    Future progress will rely on the availability of high-resolution, large-area ST datasets, enabling more rigorous real-world validation and extending our method toward predicting gene expression at near-single-cell or subcellular resolution with emerging ST technologies. \citep{Chen2021SpatiotemporalTA, poovathingal2024nova}.

\section{Conclusion}\label{conclusion}
    % In this work, we proposed PH2ST, a spatial transcriptomics-guided histological hypergraph learning framework for predicting spatial gene expression. 
    % We demonstrated the feasibility of using a small amount of ST data as a prompt to guide histological features in predicting gene expression across entire WSIs. 
    % Experiments validated the superior performance of PH2ST, highlighting its potential to enable cost-effective WSI spatial transcriptomics acquisition and advance integrative research in pathology and biomedicine.
    In this work, we introduced a novel inference-time prompting paradigm for ST prediction, aligning the task more closely with clinical practice by leveraging a small subset of known spots per WSI to guide gene expression prediction in the remaining regions. To this end, we proposed PH2ST, a novel framework that utilizes limited ST data as prompts to learn the associations between histological features and gene expression, thereby improving spatial prediction performance. 
    Central to PH2ST is the construction of dual-scale hypergraphs to effectively fuse spatial and contextual information derived from tissue morphology, as well as a cross-attention mechanism that aligns histopathological representations with ST prompts.
    Comprehensive evaluation on two publicly available ST datasets across four prompting scenarios, designed to reflect real-world needs, demonstrates that PH2ST achieves state-of-the-art performance and exhibits strong practical applicability.

    \section*{Code available}
    The code for the PH2ST is available on GitHub (\href{https://github.com/NIUYI0511/PH2ST/}{https://gi\\thub.com/NIUYI0511/PH2ST/}).
    
    \section*{Data availability}
    The human HER2-positive breast tumor spatial transcriptomics dataset is available at \href{https://github.com/almaan/her2st/}{https://github.com/almaan\\/her2st/}. The human cutaneous squamous cell carcinoma 10x Visium data is available at \href{https://www.ncbi.nlm.nih.gov/geo/query/acc.cgi?acc=GSE144240}{https://www.ncbi.nlm.nih.gov/ge\\o/query/acc.cgi?acc=GSE144240}.

	\section*{Declaration of competing interest}
	The authors declare that they have no known competing financial interests or personal relationships that could have appeared to influence the work reported in this paper.
	
    \section*{Acknowledgments}
	
	This work has been supported by the Key Research and Development Program of Ningxia Hui Nationality Autonomous Region under Grant 2022BEG02025 and 2023BE-G02023.

	\printcredits
	
	\bibliographystyle{cas-model2-names}
	
	\bibliography{PH2ST}

\begin{thebibliography}{69}
\expandafter\ifx\csname natexlab\endcsname\relax\def\natexlab#1{#1}\fi
\providecommand{\url}[1]{\texttt{#1}}
\providecommand{\href}[2]{#2}
\providecommand{\path}[1]{#1}
\providecommand{\DOIprefix}{doi:}
\providecommand{\ArXivprefix}{arXiv:}
\providecommand{\URLprefix}{URL: }
\providecommand{\Pubmedprefix}{pmid:}
\providecommand{\doi}[1]{\href{http://dx.doi.org/#1}{\path{#1}}}
\providecommand{\Pubmed}[1]{\href{pmid:#1}{\path{#1}}}
\providecommand{\bibinfo}[2]{#2}
\ifx\xfnm\relax \def\xfnm[#1]{\unskip,\space#1}\fi
%Type = Article
\bibitem[{Agarap(2018)}]{agarap2018deep}
\bibinfo{author}{Agarap, A.F.}, \bibinfo{year}{2018}.
\newblock \bibinfo{title}{Deep learning using rectified linear units (relu)}.
\newblock \bibinfo{journal}{arXiv preprint arXiv:1803.08375} .
%Type = Article
\bibitem[{Ahmedt-Aristizabal et~al.(2021)Ahmedt-Aristizabal, Armin, Denman, Fookes and Petersson}]{ahmedt2021graph}
\bibinfo{author}{Ahmedt-Aristizabal, D.}, \bibinfo{author}{Armin, M.A.}, \bibinfo{author}{Denman, S.}, \bibinfo{author}{Fookes, C.}, \bibinfo{author}{Petersson, L.}, \bibinfo{year}{2021}.
\newblock \bibinfo{title}{Graph-based deep learning for medical diagnosis and analysis: past, present and future}.
\newblock \bibinfo{journal}{Sensors} \bibinfo{volume}{21}, \bibinfo{pages}{4758}.
%Type = Article
\bibitem[{Alexey(2020)}]{alexey2020image}
\bibinfo{author}{Alexey, D.}, \bibinfo{year}{2020}.
\newblock \bibinfo{title}{An image is worth 16x16 words: Transformers for image recognition at scale}.
\newblock \bibinfo{journal}{arXiv preprint arXiv: 2010.11929} .
%Type = Article
\bibitem[{Alfonzo et~al.(2021)Alfonzo, Brown, Byers, Cheung, Maraia and Ross}]{alfonzo2021call}
\bibinfo{author}{Alfonzo, J.D.}, \bibinfo{author}{Brown, J.A.}, \bibinfo{author}{Byers, P.H.}, \bibinfo{author}{Cheung, V.G.}, \bibinfo{author}{Maraia, R.J.}, \bibinfo{author}{Ross, R.L.}, \bibinfo{year}{2021}.
\newblock \bibinfo{title}{A call for direct sequencing of full-length rnas to identify all modifications}.
\newblock \bibinfo{journal}{Nature genetics} \bibinfo{volume}{53}, \bibinfo{pages}{1113--1116}.
%Type = Book
\bibitem[{Amin et~al.(2017)Amin, Edge, Greene, Byrd, Brookland, Washington, Gershenwald, Compton, Hess, Sullivan et~al.}]{amin2017ajcc}
\bibinfo{author}{Amin, M.B.}, \bibinfo{author}{Edge, S.B.}, \bibinfo{author}{Greene, F.L.}, \bibinfo{author}{Byrd, D.R.}, \bibinfo{author}{Brookland, R.K.}, \bibinfo{author}{Washington, M.K.}, \bibinfo{author}{Gershenwald, J.E.}, \bibinfo{author}{Compton, C.C.}, \bibinfo{author}{Hess, K.R.}, \bibinfo{author}{Sullivan, D.C.}, et~al., \bibinfo{year}{2017}.
\newblock \bibinfo{title}{AJCC cancer staging manual}. volume \bibinfo{volume}{1024}.
\newblock \bibinfo{publisher}{Springer}.
%Type = Article
\bibitem[{Andersson et~al.(2021)Andersson, Larsson, Stenbeck, Salm{\'e}n, Ehinger, Wu, Al-Eryani, Roden, Swarbrick, Borg et~al.}]{andersson2021spatial}
\bibinfo{author}{Andersson, A.}, \bibinfo{author}{Larsson, L.}, \bibinfo{author}{Stenbeck, L.}, \bibinfo{author}{Salm{\'e}n, F.}, \bibinfo{author}{Ehinger, A.}, \bibinfo{author}{Wu, S.Z.}, \bibinfo{author}{Al-Eryani, G.}, \bibinfo{author}{Roden, D.}, \bibinfo{author}{Swarbrick, A.}, \bibinfo{author}{Borg, {\AA}.}, et~al., \bibinfo{year}{2021}.
\newblock \bibinfo{title}{Spatial deconvolution of her2-positive breast cancer delineates tumor-associated cell type interactions}.
\newblock \bibinfo{journal}{Nature communications} \bibinfo{volume}{12}, \bibinfo{pages}{6012}.
%Type = Article
\bibitem[{Bai et~al.(2021)Bai, Zhang and Torr}]{bai2021hypergraph}
\bibinfo{author}{Bai, S.}, \bibinfo{author}{Zhang, F.}, \bibinfo{author}{Torr, P.H.}, \bibinfo{year}{2021}.
\newblock \bibinfo{title}{Hypergraph convolution and hypergraph attention}.
\newblock \bibinfo{journal}{Pattern Recognition} \bibinfo{volume}{110}, \bibinfo{pages}{107637}.
%Type = Article
\bibitem[{Barabasi and Oltvai(2004)}]{barabasi2004network}
\bibinfo{author}{Barabasi, A.L.}, \bibinfo{author}{Oltvai, Z.N.}, \bibinfo{year}{2004}.
\newblock \bibinfo{title}{Network biology: understanding the cell's functional organization}.
\newblock \bibinfo{journal}{Nature reviews genetics} \bibinfo{volume}{5}, \bibinfo{pages}{101--113}.
%Type = Article
\bibitem[{Burgess(2019)}]{burgess2019spatial}
\bibinfo{author}{Burgess, D.J.}, \bibinfo{year}{2019}.
\newblock \bibinfo{title}{Spatial transcriptomics coming of age}.
\newblock \bibinfo{journal}{Nature Reviews Genetics} \bibinfo{volume}{20}, \bibinfo{pages}{317--317}.
%Type = Article
\bibitem[{Chen et~al.(2022)Chen, Liao, Cheng, Ma, Wu, Lai, Qiu, Yang, Xu, Hao et~al.}]{Chen2021SpatiotemporalTA}
\bibinfo{author}{Chen, A.}, \bibinfo{author}{Liao, S.}, \bibinfo{author}{Cheng, M.}, \bibinfo{author}{Ma, K.}, \bibinfo{author}{Wu, L.}, \bibinfo{author}{Lai, Y.}, \bibinfo{author}{Qiu, X.}, \bibinfo{author}{Yang, J.}, \bibinfo{author}{Xu, J.}, \bibinfo{author}{Hao, S.}, et~al., \bibinfo{year}{2022}.
\newblock \bibinfo{title}{Spatiotemporal transcriptomic atlas of mouse organogenesis using dna nanoball-patterned arrays}.
\newblock \bibinfo{journal}{Cell} \bibinfo{volume}{185}, \bibinfo{pages}{1777--1792}.
%Type = Article
\bibitem[{Chen et~al.(2024a)Chen, Zhang, Tang, Liu and Huang}]{2024Edge}
\bibinfo{author}{Chen, C.}, \bibinfo{author}{Zhang, Z.}, \bibinfo{author}{Tang, P.}, \bibinfo{author}{Liu, X.}, \bibinfo{author}{Huang, B.}, \bibinfo{year}{2024}a.
\newblock \bibinfo{title}{Edge-relational window-attentional graph neural network for gene expression prediction in spatial transcriptomics analysis}.
\newblock \bibinfo{journal}{Computers in Biology and Medicine} \bibinfo{volume}{174}.
%Type = Article
\bibitem[{Chen et~al.(2024b)Chen, Ding, Lu, Williamson, Jaume, Chen, Zhang, Shao, Song, Shaban et~al.}]{chen2024uni}
\bibinfo{author}{Chen, R.J.}, \bibinfo{author}{Ding, T.}, \bibinfo{author}{Lu, M.Y.}, \bibinfo{author}{Williamson, D.F.}, \bibinfo{author}{Jaume, G.}, \bibinfo{author}{Chen, B.}, \bibinfo{author}{Zhang, A.}, \bibinfo{author}{Shao, D.}, \bibinfo{author}{Song, A.H.}, \bibinfo{author}{Shaban, M.}, et~al., \bibinfo{year}{2024}b.
\newblock \bibinfo{title}{Towards a general-purpose foundation model for computational pathology}.
\newblock \bibinfo{journal}{Nature Medicine} .
%Type = Incollection
\bibitem[{Chen et~al.(2021)Chen, Lu, Shaban, Chen, Chen, Williamson and Mahmood}]{chen2021whole}
\bibinfo{author}{Chen, R.J.}, \bibinfo{author}{Lu, M.Y.}, \bibinfo{author}{Shaban, M.}, \bibinfo{author}{Chen, C.}, \bibinfo{author}{Chen, T.Y.}, \bibinfo{author}{Williamson, D.F.}, \bibinfo{author}{Mahmood, F.}, \bibinfo{year}{2021}.
\newblock \bibinfo{title}{Whole slide images are 2d point clouds: Context-aware survival prediction using patch-based graph convolutional networks}, in: \bibinfo{booktitle}{Medical Image Computing and Computer Assisted Intervention {\textendash} {MICCAI} 2021}. \bibinfo{publisher}{Springer International Publishing}, pp. \bibinfo{pages}{339--349}.
%Type = Misc
\bibitem[{Cheung(2018)}]{cheung2018classification}
\bibinfo{author}{Cheung, A.}, \bibinfo{year}{2018}.
\newblock \bibinfo{title}{Who classification of tumours 5th edition}.
%Type = Inproceedings
\bibitem[{Chung et~al.(2024)Chung, Ha, Im and Lee}]{chung2024accurate}
\bibinfo{author}{Chung, Y.}, \bibinfo{author}{Ha, J.H.}, \bibinfo{author}{Im, K.C.}, \bibinfo{author}{Lee, J.S.}, \bibinfo{year}{2024}.
\newblock \bibinfo{title}{Accurate spatial gene expression prediction by integrating multi-resolution features}, in: \bibinfo{booktitle}{Proceedings of the IEEE/CVF Conference on Computer Vision and Pattern Recognition}, pp. \bibinfo{pages}{11591--11600}.
%Type = Article
\bibitem[{Consortium et~al.(2010)}]{10002010map}
\bibinfo{author}{Consortium, .G.P.}, et~al., \bibinfo{year}{2010}.
\newblock \bibinfo{title}{A map of human genome variation from population scale sequencing}.
\newblock \bibinfo{journal}{Nature} \bibinfo{volume}{467}, \bibinfo{pages}{1061}.
%Type = Article
\bibitem[{Cui~Zhou et~al.(2022)Cui~Zhou, Jayasinghe, Chen, Herndon, Iglesia, Navale, Wendl, Caravan, Sato, Storrs et~al.}]{cui2022spatially}
\bibinfo{author}{Cui~Zhou, D.}, \bibinfo{author}{Jayasinghe, R.G.}, \bibinfo{author}{Chen, S.}, \bibinfo{author}{Herndon, J.M.}, \bibinfo{author}{Iglesia, M.D.}, \bibinfo{author}{Navale, P.}, \bibinfo{author}{Wendl, M.C.}, \bibinfo{author}{Caravan, W.}, \bibinfo{author}{Sato, K.}, \bibinfo{author}{Storrs, E.}, et~al., \bibinfo{year}{2022}.
\newblock \bibinfo{title}{Spatially restricted drivers and transitional cell populations cooperate with the microenvironment in untreated and chemo-resistant pancreatic cancer}.
\newblock \bibinfo{journal}{Nature genetics} \bibinfo{volume}{54}, \bibinfo{pages}{1390--1405}.
%Type = Article
\bibitem[{El~Nahhas et~al.(2024)El~Nahhas, Loeffler, Carrero, van Treeck, Kolbinger, Hewitt, Muti, Graziani, Zeng, Calderaro et~al.}]{el2024regression}
\bibinfo{author}{El~Nahhas, O.S.}, \bibinfo{author}{Loeffler, C.M.}, \bibinfo{author}{Carrero, Z.I.}, \bibinfo{author}{van Treeck, M.}, \bibinfo{author}{Kolbinger, F.R.}, \bibinfo{author}{Hewitt, K.J.}, \bibinfo{author}{Muti, H.S.}, \bibinfo{author}{Graziani, M.}, \bibinfo{author}{Zeng, Q.}, \bibinfo{author}{Calderaro, J.}, et~al., \bibinfo{year}{2024}.
\newblock \bibinfo{title}{Regression-based deep-learning predicts molecular biomarkers from pathology slides}.
\newblock \bibinfo{journal}{nature communications} \bibinfo{volume}{15}, \bibinfo{pages}{1253}.
%Type = Inproceedings
\bibitem[{Feng et~al.(2019)Feng, You, Zhang, Ji and Gao}]{feng2019hypergraph}
\bibinfo{author}{Feng, Y.}, \bibinfo{author}{You, H.}, \bibinfo{author}{Zhang, Z.}, \bibinfo{author}{Ji, R.}, \bibinfo{author}{Gao, Y.}, \bibinfo{year}{2019}.
\newblock \bibinfo{title}{Hypergraph neural networks}, in: \bibinfo{booktitle}{Proceedings of the AAAI conference on artificial intelligence}, pp. \bibinfo{pages}{3558--3565}.
%Type = Article
\bibitem[{Garraway and Lander(2013)}]{garraway2013lessons}
\bibinfo{author}{Garraway, L.A.}, \bibinfo{author}{Lander, E.S.}, \bibinfo{year}{2013}.
\newblock \bibinfo{title}{Lessons from the cancer genome}.
\newblock \bibinfo{journal}{Cell} \bibinfo{volume}{153}, \bibinfo{pages}{17--37}.
%Type = Article
\bibitem[{He et~al.(2020)He, Bergenstr{\aa}hle, Stenbeck, Abid, Andersson, Borg, Maaskola, Lundeberg and Zou}]{he2020integrating}
\bibinfo{author}{He, B.}, \bibinfo{author}{Bergenstr{\aa}hle, L.}, \bibinfo{author}{Stenbeck, L.}, \bibinfo{author}{Abid, A.}, \bibinfo{author}{Andersson, A.}, \bibinfo{author}{Borg, {\AA}.}, \bibinfo{author}{Maaskola, J.}, \bibinfo{author}{Lundeberg, J.}, \bibinfo{author}{Zou, J.}, \bibinfo{year}{2020}.
\newblock \bibinfo{title}{Integrating spatial gene expression and breast tumour morphology via deep learning}.
\newblock \bibinfo{journal}{Nature biomedical engineering} \bibinfo{volume}{4}, \bibinfo{pages}{827--834}.
%Type = Article
\bibitem[{He et~al.(2016)He, Zhang, Ren and Sun}]{2016Deep}
\bibinfo{author}{He, K.}, \bibinfo{author}{Zhang, X.}, \bibinfo{author}{Ren, S.}, \bibinfo{author}{Sun, J.}, \bibinfo{year}{2016}.
\newblock \bibinfo{title}{Deep residual learning for image recognition}.
\newblock \bibinfo{journal}{IEEE} .
%Type = Article
\bibitem[{Hendrycks and Gimpel(2016)}]{hendrycks2016gaussian}
\bibinfo{author}{Hendrycks, D.}, \bibinfo{author}{Gimpel, K.}, \bibinfo{year}{2016}.
\newblock \bibinfo{title}{Gaussian error linear units (gelus)}.
\newblock \bibinfo{journal}{arXiv preprint arXiv:1606.08415} .
%Type = Article
\bibitem[{Hoang et~al.(2024)Hoang, Shulman, Turakulov, Abdullaev, Singh, Campagnolo, Lalchungnunga, Stone, Nasrallah, Ruppin et~al.}]{hoang2024prediction}
\bibinfo{author}{Hoang, D.T.}, \bibinfo{author}{Shulman, E.D.}, \bibinfo{author}{Turakulov, R.}, \bibinfo{author}{Abdullaev, Z.}, \bibinfo{author}{Singh, O.}, \bibinfo{author}{Campagnolo, E.M.}, \bibinfo{author}{Lalchungnunga, H.}, \bibinfo{author}{Stone, E.A.}, \bibinfo{author}{Nasrallah, M.P.}, \bibinfo{author}{Ruppin, E.}, et~al., \bibinfo{year}{2024}.
\newblock \bibinfo{title}{Prediction of dna methylation-based tumor types from histopathology in central nervous system tumors with deep learning}.
\newblock \bibinfo{journal}{Nature Medicine} \bibinfo{volume}{30}, \bibinfo{pages}{1952--1961}.
%Type = Article
\bibitem[{Hu et~al.(2021)Hu, Li, Coleman, Schroeder, Ma, Irwin, Lee, Shinohara and Li}]{hu2021spagcn}
\bibinfo{author}{Hu, J.}, \bibinfo{author}{Li, X.}, \bibinfo{author}{Coleman, K.}, \bibinfo{author}{Schroeder, A.}, \bibinfo{author}{Ma, N.}, \bibinfo{author}{Irwin, D.J.}, \bibinfo{author}{Lee, E.B.}, \bibinfo{author}{Shinohara, R.T.}, \bibinfo{author}{Li, M.}, \bibinfo{year}{2021}.
\newblock \bibinfo{title}{Spagcn: Integrating gene expression, spatial location and histology to identify spatial domains and spatially variable genes by graph convolutional network}.
\newblock \bibinfo{journal}{Nature methods} \bibinfo{volume}{18}, \bibinfo{pages}{1342--1351}.
%Type = Inproceedings
\bibitem[{Huang et~al.(2017)Huang, Liu, Van Der~Maaten and Weinberger}]{huang2017densely}
\bibinfo{author}{Huang, G.}, \bibinfo{author}{Liu, Z.}, \bibinfo{author}{Van Der~Maaten, L.}, \bibinfo{author}{Weinberger, K.Q.}, \bibinfo{year}{2017}.
\newblock \bibinfo{title}{Densely connected convolutional networks}, in: \bibinfo{booktitle}{Proceedings of the IEEE conference on computer vision and pattern recognition}, pp. \bibinfo{pages}{4700--4708}.
%Type = Article
\bibitem[{Jain and Eadon(2024)}]{jain2024spatial}
\bibinfo{author}{Jain, S.}, \bibinfo{author}{Eadon, M.T.}, \bibinfo{year}{2024}.
\newblock \bibinfo{title}{Spatial transcriptomics in health and disease}.
\newblock \bibinfo{journal}{Nature Reviews Nephrology} \bibinfo{volume}{20}, \bibinfo{pages}{659--671}.
%Type = Article
\bibitem[{Janesick et~al.(2022)Janesick, Shelansky, Gottscho, Wagner, Rouault, Beliakoff, de~Oliveira, Kohlway, Abousoud, Morrison et~al.}]{janesick2022high}
\bibinfo{author}{Janesick, A.}, \bibinfo{author}{Shelansky, R.}, \bibinfo{author}{Gottscho, A.}, \bibinfo{author}{Wagner, F.}, \bibinfo{author}{Rouault, M.}, \bibinfo{author}{Beliakoff, G.}, \bibinfo{author}{de~Oliveira, M.F.}, \bibinfo{author}{Kohlway, A.}, \bibinfo{author}{Abousoud, J.}, \bibinfo{author}{Morrison, C.}, et~al., \bibinfo{year}{2022}.
\newblock \bibinfo{title}{High resolution mapping of the breast cancer tumor microenvironment using integrated single cell, spatial and in situ analysis of ffpe tissue}.
\newblock \bibinfo{journal}{Biorxiv} , \bibinfo{pages}{2022--10}.
%Type = Inproceedings
\bibitem[{Jaume et~al.(2024)Jaume, Doucet, Song, Lu, Almagro-Perez, Wagner, Vaidya, Chen, Williamson, Kim and Mahmood}]{jaume2024hest}
\bibinfo{author}{Jaume, G.}, \bibinfo{author}{Doucet, P.}, \bibinfo{author}{Song, A.H.}, \bibinfo{author}{Lu, M.Y.}, \bibinfo{author}{Almagro-Perez, C.}, \bibinfo{author}{Wagner, S.J.}, \bibinfo{author}{Vaidya, A.J.}, \bibinfo{author}{Chen, R.J.}, \bibinfo{author}{Williamson, D.F.K.}, \bibinfo{author}{Kim, A.}, \bibinfo{author}{Mahmood, F.}, \bibinfo{year}{2024}.
\newblock \bibinfo{title}{Hest-1k: A dataset for spatial transcriptomics and histology image analysis}, in: \bibinfo{booktitle}{Advances in Neural Information Processing Systems}.
%Type = Article
\bibitem[{Ji et~al.(2020)Ji, Rubin, Thrane, Jiang, Reynolds, Meyers, Guo, George, Mollbrink, Bergenstr{\aa}hle et~al.}]{ji2020multimodal}
\bibinfo{author}{Ji, A.L.}, \bibinfo{author}{Rubin, A.J.}, \bibinfo{author}{Thrane, K.}, \bibinfo{author}{Jiang, S.}, \bibinfo{author}{Reynolds, D.L.}, \bibinfo{author}{Meyers, R.M.}, \bibinfo{author}{Guo, M.G.}, \bibinfo{author}{George, B.M.}, \bibinfo{author}{Mollbrink, A.}, \bibinfo{author}{Bergenstr{\aa}hle, J.}, et~al., \bibinfo{year}{2020}.
\newblock \bibinfo{title}{Multimodal analysis of composition and spatial architecture in human squamous cell carcinoma}.
\newblock \bibinfo{journal}{Cell} \bibinfo{volume}{182}, \bibinfo{pages}{497--514}.
%Type = Article
\bibitem[{Jia et~al.(2024)Jia, Liu, Chen, Zhao and Wang}]{jia2024thitogene}
\bibinfo{author}{Jia, Y.}, \bibinfo{author}{Liu, J.}, \bibinfo{author}{Chen, L.}, \bibinfo{author}{Zhao, T.}, \bibinfo{author}{Wang, Y.}, \bibinfo{year}{2024}.
\newblock \bibinfo{title}{Thitogene: a deep learning method for predicting spatial transcriptomics from histological images}.
\newblock \bibinfo{journal}{Briefings in Bioinformatics} \bibinfo{volume}{25}, \bibinfo{pages}{bbad464}.
%Type = Article
\bibitem[{Kharchenko et~al.(2014)Kharchenko, Silberstein and Scadden}]{kharchenko2014bayesian}
\bibinfo{author}{Kharchenko, P.V.}, \bibinfo{author}{Silberstein, L.}, \bibinfo{author}{Scadden, D.T.}, \bibinfo{year}{2014}.
\newblock \bibinfo{title}{Bayesian approach to single-cell differential expression analysis}.
\newblock \bibinfo{journal}{Nature methods} \bibinfo{volume}{11}, \bibinfo{pages}{740--742}.
%Type = Article
\bibitem[{Kingma(2014)}]{kingma2014adam}
\bibinfo{author}{Kingma, D.P.}, \bibinfo{year}{2014}.
\newblock \bibinfo{title}{Adam: A method for stochastic optimization}.
\newblock \bibinfo{journal}{arXiv preprint arXiv:1412.6980} .
%Type = Article
\bibitem[{Lee and Jang(2022)}]{lee2022deep}
\bibinfo{author}{Lee, S.H.}, \bibinfo{author}{Jang, H.J.}, \bibinfo{year}{2022}.
\newblock \bibinfo{title}{Deep learning-based prediction of molecular cancer biomarkers from tissue slides: A new tool for precision oncology}.
\newblock \bibinfo{journal}{Clinical and Molecular Hepatology} \bibinfo{volume}{28}, \bibinfo{pages}{754}.
%Type = Article
\bibitem[{Li et~al.(2024a)Li, Bao, Hou, Li, Li, Deng and Dai}]{li2024tissue}
\bibinfo{author}{Li, B.}, \bibinfo{author}{Bao, F.}, \bibinfo{author}{Hou, Y.}, \bibinfo{author}{Li, F.}, \bibinfo{author}{Li, H.}, \bibinfo{author}{Deng, Y.}, \bibinfo{author}{Dai, Q.}, \bibinfo{year}{2024}a.
\newblock \bibinfo{title}{Tissue characterization at an enhanced resolution across spatial omics platforms with deep generative model}.
\newblock \bibinfo{journal}{Nature Communications} \bibinfo{volume}{15}, \bibinfo{pages}{6541}.
%Type = Inproceedings
\bibitem[{Li et~al.(2021)Li, Li and Eliceiri}]{li2021dual}
\bibinfo{author}{Li, B.}, \bibinfo{author}{Li, Y.}, \bibinfo{author}{Eliceiri, K.W.}, \bibinfo{year}{2021}.
\newblock \bibinfo{title}{Dual-stream multiple instance learning network for whole slide image classification with self-supervised contrastive learning}, in: \bibinfo{booktitle}{Proceedings of the IEEE/CVF conference on computer vision and pattern recognition}, pp. \bibinfo{pages}{14318--14328}.
%Type = Article
\bibitem[{Li et~al.(2024b)Li, Zhang, Wang, Zhang, Li, Wang and Song}]{li2024gene}
\bibinfo{author}{Li, B.}, \bibinfo{author}{Zhang, Y.}, \bibinfo{author}{Wang, Q.}, \bibinfo{author}{Zhang, C.}, \bibinfo{author}{Li, M.}, \bibinfo{author}{Wang, G.}, \bibinfo{author}{Song, Q.}, \bibinfo{year}{2024}b.
\newblock \bibinfo{title}{Gene expression prediction from histology images via hypergraph neural networks}.
\newblock \bibinfo{journal}{Briefings in Bioinformatics} \bibinfo{volume}{25}, \bibinfo{pages}{bbae500}.
%Type = Article
\bibitem[{Liu et~al.(2023)Liu, Zhou and Lei}]{liu2023rmdgcn}
\bibinfo{author}{Liu, L.}, \bibinfo{author}{Zhou, Y.}, \bibinfo{author}{Lei, X.}, \bibinfo{year}{2023}.
\newblock \bibinfo{title}{Rmdgcn: Prediction of rna methylation and disease associations based on graph convolutional network with attention mechanism}.
\newblock \bibinfo{journal}{PLoS Computational Biology} \bibinfo{volume}{19}, \bibinfo{pages}{e1011677}.
%Type = Article
\bibitem[{Lopez et~al.(2019)Lopez, Nazaret, Langevin, Samaran, Regier, Jordan and Yosef}]{lopez2019joint}
\bibinfo{author}{Lopez, R.}, \bibinfo{author}{Nazaret, A.}, \bibinfo{author}{Langevin, M.}, \bibinfo{author}{Samaran, J.}, \bibinfo{author}{Regier, J.}, \bibinfo{author}{Jordan, M.I.}, \bibinfo{author}{Yosef, N.}, \bibinfo{year}{2019}.
\newblock \bibinfo{title}{A joint model of unpaired data from scrna-seq and spatial transcriptomics for imputing missing gene expression measurements}.
\newblock \bibinfo{journal}{arXiv preprint arXiv:1905.02269} .
%Type = Article
\bibitem[{Lu et~al.(2024)Lu, Chen, Williamson, Chen, Liang, Ding, Jaume, Odintsov, Le, Gerber et~al.}]{lu2024avisionlanguage}
\bibinfo{author}{Lu, M.Y.}, \bibinfo{author}{Chen, B.}, \bibinfo{author}{Williamson, D.F.}, \bibinfo{author}{Chen, R.J.}, \bibinfo{author}{Liang, I.}, \bibinfo{author}{Ding, T.}, \bibinfo{author}{Jaume, G.}, \bibinfo{author}{Odintsov, I.}, \bibinfo{author}{Le, L.P.}, \bibinfo{author}{Gerber, G.}, et~al., \bibinfo{year}{2024}.
\newblock \bibinfo{title}{A visual-language foundation model for computational pathology}.
\newblock \bibinfo{journal}{Nature Medicine} \bibinfo{volume}{30}, \bibinfo{pages}{863–874}.
%Type = Article
\bibitem[{Lu et~al.(2021)Lu, Williamson, Chen, Chen, Barbieri and Mahmood}]{lu2021data}
\bibinfo{author}{Lu, M.Y.}, \bibinfo{author}{Williamson, D.F.}, \bibinfo{author}{Chen, T.Y.}, \bibinfo{author}{Chen, R.J.}, \bibinfo{author}{Barbieri, M.}, \bibinfo{author}{Mahmood, F.}, \bibinfo{year}{2021}.
\newblock \bibinfo{title}{Data-efficient and weakly supervised computational pathology on whole-slide images}.
\newblock \bibinfo{journal}{Nature biomedical engineering} \bibinfo{volume}{5}, \bibinfo{pages}{555--570}.
%Type = Article
\bibitem[{Moffitt et~al.(2022)Moffitt, Lundberg and Heyn}]{Moffitt2022the}
\bibinfo{author}{Moffitt, J.R.}, \bibinfo{author}{Lundberg, E.}, \bibinfo{author}{Heyn, H.}, \bibinfo{year}{2022}.
\newblock \bibinfo{title}{The emerging landscape of spatial profiling technologies}.
\newblock \bibinfo{journal}{Nature reviews. Genetics} \bibinfo{volume}{23}, \bibinfo{pages}{741—759}.
\newblock \DOIprefix\doi{10.1038/s41576-022-00515-3}.
%Type = Article
\bibitem[{Moncada et~al.(2020)Moncada, Barkley, Wagner, Chiodin, Devlin, Baron, Hajdu, Simeone and Yanai}]{moncada2020integrating}
\bibinfo{author}{Moncada, R.}, \bibinfo{author}{Barkley, D.}, \bibinfo{author}{Wagner, F.}, \bibinfo{author}{Chiodin, M.}, \bibinfo{author}{Devlin, J.C.}, \bibinfo{author}{Baron, M.}, \bibinfo{author}{Hajdu, C.H.}, \bibinfo{author}{Simeone, D.M.}, \bibinfo{author}{Yanai, I.}, \bibinfo{year}{2020}.
\newblock \bibinfo{title}{Integrating microarray-based spatial transcriptomics and single-cell rna-seq reveals tissue architecture in pancreatic ductal adenocarcinomas}.
\newblock \bibinfo{journal}{Nature biotechnology} \bibinfo{volume}{38}, \bibinfo{pages}{333--342}.
%Type = Article
\bibitem[{Ober and Vercelli(2011)}]{ober2011gene}
\bibinfo{author}{Ober, C.}, \bibinfo{author}{Vercelli, D.}, \bibinfo{year}{2011}.
\newblock \bibinfo{title}{Gene--environment interactions in human disease: nuisance or opportunity?}
\newblock \bibinfo{journal}{Trends in genetics} \bibinfo{volume}{27}, \bibinfo{pages}{107--115}.
%Type = Article
\bibitem[{Palla et~al.(2022)Palla, Fischer, Regev and Theis}]{palla2022spatial}
\bibinfo{author}{Palla, G.}, \bibinfo{author}{Fischer, D.S.}, \bibinfo{author}{Regev, A.}, \bibinfo{author}{Theis, F.J.}, \bibinfo{year}{2022}.
\newblock \bibinfo{title}{Spatial components of molecular tissue biology}.
\newblock \bibinfo{journal}{Nature Biotechnology} \bibinfo{volume}{40}, \bibinfo{pages}{308--318}.
%Type = Article
\bibitem[{Pang et~al.(2021)Pang, Su and Li}]{pang2021leveraging}
\bibinfo{author}{Pang, M.}, \bibinfo{author}{Su, K.}, \bibinfo{author}{Li, M.}, \bibinfo{year}{2021}.
\newblock \bibinfo{title}{Leveraging information in spatial transcriptomics to predict super-resolution gene expression from histology images in tumors}.
\newblock \bibinfo{journal}{BioRxiv} , \bibinfo{pages}{2021--11}.
%Type = Article
\bibitem[{Papanicolau-Sengos and Aldape(2022)}]{papanicolau2022dna}
\bibinfo{author}{Papanicolau-Sengos, A.}, \bibinfo{author}{Aldape, K.}, \bibinfo{year}{2022}.
\newblock \bibinfo{title}{Dna methylation profiling: an emerging paradigm for cancer diagnosis}.
\newblock \bibinfo{journal}{Annual Review of Pathology: Mechanisms of Disease} \bibinfo{volume}{17}, \bibinfo{pages}{295--321}.
%Type = Article
\bibitem[{Pizurica et~al.(2024)Pizurica, Zheng, Carrillo-Perez, Noor, Yao, Wohlfart, Vladimirova, Marchal and Gevaert}]{pizurica2024digital}
\bibinfo{author}{Pizurica, M.}, \bibinfo{author}{Zheng, Y.}, \bibinfo{author}{Carrillo-Perez, F.}, \bibinfo{author}{Noor, H.}, \bibinfo{author}{Yao, W.}, \bibinfo{author}{Wohlfart, C.}, \bibinfo{author}{Vladimirova, A.}, \bibinfo{author}{Marchal, K.}, \bibinfo{author}{Gevaert, O.}, \bibinfo{year}{2024}.
\newblock \bibinfo{title}{Digital profiling of gene expression from histology images with linearized attention}.
\newblock \bibinfo{journal}{Nature Communications} \bibinfo{volume}{15}, \bibinfo{pages}{9886}.
%Type = Article
\bibitem[{Poovathingal et~al.(2024)Poovathingal, Davie, Borm, Vandepoel, Poulvellarie, Verfaillie, Corthout and Aerts}]{poovathingal2024nova}
\bibinfo{author}{Poovathingal, S.}, \bibinfo{author}{Davie, K.}, \bibinfo{author}{Borm, L.E.}, \bibinfo{author}{Vandepoel, R.}, \bibinfo{author}{Poulvellarie, N.}, \bibinfo{author}{Verfaillie, A.}, \bibinfo{author}{Corthout, N.}, \bibinfo{author}{Aerts, S.}, \bibinfo{year}{2024}.
\newblock \bibinfo{title}{Nova-st: Nano-patterned ultra-dense platform for spatial transcriptomics}.
\newblock \bibinfo{journal}{Cell Reports Methods} \bibinfo{volume}{4}.
%Type = Article
\bibitem[{Rao et~al.(2021a)Rao, Barkley, Fran{\c{c}}a and Yanai}]{rao2021exploring}
\bibinfo{author}{Rao, A.}, \bibinfo{author}{Barkley, D.}, \bibinfo{author}{Fran{\c{c}}a, G.S.}, \bibinfo{author}{Yanai, I.}, \bibinfo{year}{2021}a.
\newblock \bibinfo{title}{Exploring tissue architecture using spatial transcriptomics}.
\newblock \bibinfo{journal}{Nature} \bibinfo{volume}{596}, \bibinfo{pages}{211--220}.
%Type = Article
\bibitem[{Rao et~al.(2021b)Rao, Zhou, Lu, Zhao and Yang}]{rao2021imputing}
\bibinfo{author}{Rao, J.}, \bibinfo{author}{Zhou, X.}, \bibinfo{author}{Lu, Y.}, \bibinfo{author}{Zhao, H.}, \bibinfo{author}{Yang, Y.}, \bibinfo{year}{2021}b.
\newblock \bibinfo{title}{Imputing single-cell rna-seq data by combining graph convolution and autoencoder neural networks}.
\newblock \bibinfo{journal}{Iscience} \bibinfo{volume}{24}.
%Type = Article
\bibitem[{Schapke et~al.(2021)Schapke, Tavares and Recamonde-Mendoza}]{schapke2021epgat}
\bibinfo{author}{Schapke, J.}, \bibinfo{author}{Tavares, A.}, \bibinfo{author}{Recamonde-Mendoza, M.}, \bibinfo{year}{2021}.
\newblock \bibinfo{title}{Epgat: gene essentiality prediction with graph attention networks}.
\newblock \bibinfo{journal}{IEEE/ACM Transactions on Computational Biology and Bioinformatics} \bibinfo{volume}{19}, \bibinfo{pages}{1615--1626}.
%Type = Article
\bibitem[{Schnitt(2010)}]{schnitt2010classification}
\bibinfo{author}{Schnitt, S.J.}, \bibinfo{year}{2010}.
\newblock \bibinfo{title}{Classification and prognosis of invasive breast cancer: from morphology to molecular taxonomy}.
\newblock \bibinfo{journal}{Modern pathology} \bibinfo{volume}{23}, \bibinfo{pages}{S60--S64}.
%Type = Article
\bibitem[{Shao et~al.(2021)Shao, Bian, Chen, Wang, Zhang, Ji et~al.}]{shao2021transmil}
\bibinfo{author}{Shao, Z.}, \bibinfo{author}{Bian, H.}, \bibinfo{author}{Chen, Y.}, \bibinfo{author}{Wang, Y.}, \bibinfo{author}{Zhang, J.}, \bibinfo{author}{Ji, X.}, et~al., \bibinfo{year}{2021}.
\newblock \bibinfo{title}{Transmil: Transformer based correlated multiple instance learning for whole slide image classification}.
\newblock \bibinfo{journal}{Advances in neural information processing systems} \bibinfo{volume}{34}, \bibinfo{pages}{2136--2147}.
%Type = Article
\bibitem[{St{\aa}hl et~al.(2016)St{\aa}hl, Salm{\'e}n, Vickovic, Lundmark, Navarro, Magnusson, Giacomello, Asp, Westholm, Huss et~al.}]{staahl2016visualization}
\bibinfo{author}{St{\aa}hl, P.L.}, \bibinfo{author}{Salm{\'e}n, F.}, \bibinfo{author}{Vickovic, S.}, \bibinfo{author}{Lundmark, A.}, \bibinfo{author}{Navarro, J.F.}, \bibinfo{author}{Magnusson, J.}, \bibinfo{author}{Giacomello, S.}, \bibinfo{author}{Asp, M.}, \bibinfo{author}{Westholm, J.O.}, \bibinfo{author}{Huss, M.}, et~al., \bibinfo{year}{2016}.
\newblock \bibinfo{title}{Visualization and analysis of gene expression in tissue sections by spatial transcriptomics}.
\newblock \bibinfo{journal}{Science} \bibinfo{volume}{353}, \bibinfo{pages}{78--82}.
%Type = Article
\bibitem[{Tian et~al.(2023)Tian, Chen and Macosko}]{tian2023expanding}
\bibinfo{author}{Tian, L.}, \bibinfo{author}{Chen, F.}, \bibinfo{author}{Macosko, E.Z.}, \bibinfo{year}{2023}.
\newblock \bibinfo{title}{The expanding vistas of spatial transcriptomics}.
\newblock \bibinfo{journal}{Nature Biotechnology} \bibinfo{volume}{41}, \bibinfo{pages}{773--782}.
%Type = Article
\bibitem[{Wang et~al.(2022)Wang, Yang, Zhang, Wang, Zhang, Yang, Huang and Han}]{WANG2022102559}
\bibinfo{author}{Wang, X.}, \bibinfo{author}{Yang, S.}, \bibinfo{author}{Zhang, J.}, \bibinfo{author}{Wang, M.}, \bibinfo{author}{Zhang, J.}, \bibinfo{author}{Yang, W.}, \bibinfo{author}{Huang, J.}, \bibinfo{author}{Han, X.}, \bibinfo{year}{2022}.
\newblock \bibinfo{title}{Transformer-based unsupervised contrastive learning for histopathological image classification}.
\newblock \bibinfo{journal}{Medical Image Analysis} \bibinfo{volume}{81}, \bibinfo{pages}{102559}.
%Type = Article
\bibitem[{Wang et~al.(2021)Wang, Wang, Hu, Li, Fan, Otter, Sam, Gou, Hu, Kwok et~al.}]{wang2021cell}
\bibinfo{author}{Wang, Y.}, \bibinfo{author}{Wang, Y.G.}, \bibinfo{author}{Hu, C.}, \bibinfo{author}{Li, M.}, \bibinfo{author}{Fan, Y.}, \bibinfo{author}{Otter, N.}, \bibinfo{author}{Sam, I.}, \bibinfo{author}{Gou, H.}, \bibinfo{author}{Hu, Y.}, \bibinfo{author}{Kwok, T.}, et~al., \bibinfo{year}{2021}.
\newblock \bibinfo{title}{Cell graph neural networks enable the digital staging of tumor microenvironment and precise prediction of patient survival in gastric cancer}.
\newblock \bibinfo{journal}{MedRxiv} , \bibinfo{pages}{2021--09}.
%Type = Article
\bibitem[{Wang et~al.(2009)Wang, Gerstein and Snyder}]{wang2009rna}
\bibinfo{author}{Wang, Z.}, \bibinfo{author}{Gerstein, M.}, \bibinfo{author}{Snyder, M.}, \bibinfo{year}{2009}.
\newblock \bibinfo{title}{Rna-seq: a revolutionary tool for transcriptomics}.
\newblock \bibinfo{journal}{Nature reviews genetics} \bibinfo{volume}{10}, \bibinfo{pages}{57--63}.
%Type = Article
\bibitem[{Wen et~al.(2023)Wen, Tang, Jin, Ding, Liu, Dai, Shi, Shang, Liu and Xie}]{wen2023single}
\bibinfo{author}{Wen, H.}, \bibinfo{author}{Tang, W.}, \bibinfo{author}{Jin, W.}, \bibinfo{author}{Ding, J.}, \bibinfo{author}{Liu, R.}, \bibinfo{author}{Dai, X.}, \bibinfo{author}{Shi, F.}, \bibinfo{author}{Shang, L.}, \bibinfo{author}{Liu, H.}, \bibinfo{author}{Xie, Y.}, \bibinfo{year}{2023}.
\newblock \bibinfo{title}{Single cells are spatial tokens: Transformers for spatial transcriptomic data imputation}.
\newblock \bibinfo{journal}{arXiv preprint arXiv:2302.03038} .
%Type = Article
\bibitem[{Wirth et~al.(2023)Wirth, Huber, Yin, Brood, Chang, Martinez-Jimenez and Meier}]{wirth2023spatial}
\bibinfo{author}{Wirth, J.}, \bibinfo{author}{Huber, N.}, \bibinfo{author}{Yin, K.}, \bibinfo{author}{Brood, S.}, \bibinfo{author}{Chang, S.}, \bibinfo{author}{Martinez-Jimenez, C.P.}, \bibinfo{author}{Meier, M.}, \bibinfo{year}{2023}.
\newblock \bibinfo{title}{Spatial transcriptomics using multiplexed deterministic barcoding in tissue}.
\newblock \bibinfo{journal}{Nature Communications} \bibinfo{volume}{14}, \bibinfo{pages}{1523}.
%Type = Article
\bibitem[{Wu et~al.(2022)Wu, Sun, Zhang, Xie and Cui}]{wu2022graph}
\bibinfo{author}{Wu, S.}, \bibinfo{author}{Sun, F.}, \bibinfo{author}{Zhang, W.}, \bibinfo{author}{Xie, X.}, \bibinfo{author}{Cui, B.}, \bibinfo{year}{2022}.
\newblock \bibinfo{title}{Graph neural networks in recommender systems: a survey}.
\newblock \bibinfo{journal}{ACM Computing Surveys} \bibinfo{volume}{55}, \bibinfo{pages}{1--37}.
%Type = Article
\bibitem[{Xie et~al.(2024)Xie, Pang, Chung, Perciani, MacParland, Wang and Bader}]{xie2024spatially}
\bibinfo{author}{Xie, R.}, \bibinfo{author}{Pang, K.}, \bibinfo{author}{Chung, S.}, \bibinfo{author}{Perciani, C.}, \bibinfo{author}{MacParland, S.}, \bibinfo{author}{Wang, B.}, \bibinfo{author}{Bader, G.}, \bibinfo{year}{2024}.
\newblock \bibinfo{title}{Spatially resolved gene expression prediction from histology images via bi-modal contrastive learning}.
\newblock \bibinfo{journal}{Advances in Neural Information Processing Systems} \bibinfo{volume}{36}.
%Type = Article
\bibitem[{Xu et~al.(2024)Xu, Chun, Wang, Mei, Chen and Huang}]{xu2024dna}
\bibinfo{author}{Xu, J.}, \bibinfo{author}{Chun, H.}, \bibinfo{author}{Wang, L.}, \bibinfo{author}{Mei, H.}, \bibinfo{author}{Chen, S.}, \bibinfo{author}{Huang, X.}, \bibinfo{year}{2024}.
\newblock \bibinfo{title}{Dna microarray chips: Fabrication and cutting-edge applications}.
\newblock \bibinfo{journal}{Chemical Engineering Journal} , \bibinfo{pages}{155937}.
%Type = Article
\bibitem[{Yan et~al.(2023)Yan, Shen, Zhang, Xu, Wang, Li, Ren, Ye and Zhou}]{yan2023histopathological}
\bibinfo{author}{Yan, R.}, \bibinfo{author}{Shen, Y.}, \bibinfo{author}{Zhang, X.}, \bibinfo{author}{Xu, P.}, \bibinfo{author}{Wang, J.}, \bibinfo{author}{Li, J.}, \bibinfo{author}{Ren, F.}, \bibinfo{author}{Ye, D.}, \bibinfo{author}{Zhou, S.K.}, \bibinfo{year}{2023}.
\newblock \bibinfo{title}{Histopathological bladder cancer gene mutation prediction with hierarchical deep multiple-instance learning}.
\newblock \bibinfo{journal}{Medical Image Analysis} \bibinfo{volume}{87}, \bibinfo{pages}{102824}.
%Type = Article
\bibitem[{Zeng et~al.(2022)Zeng, Wei, Yu, Yin, Yuan, Li, Tang, Lu and Yang}]{zeng2022spatial}
\bibinfo{author}{Zeng, Y.}, \bibinfo{author}{Wei, Z.}, \bibinfo{author}{Yu, W.}, \bibinfo{author}{Yin, R.}, \bibinfo{author}{Yuan, Y.}, \bibinfo{author}{Li, B.}, \bibinfo{author}{Tang, Z.}, \bibinfo{author}{Lu, Y.}, \bibinfo{author}{Yang, Y.}, \bibinfo{year}{2022}.
\newblock \bibinfo{title}{Spatial transcriptomics prediction from histology jointly through transformer and graph neural networks}.
\newblock \bibinfo{journal}{Briefings in Bioinformatics} \bibinfo{volume}{23}, \bibinfo{pages}{bbac297}.
%Type = Article
\bibitem[{Zhang et~al.(2022)Zhang, Chen, Song, Liu, Zhang, Xu and Wang}]{zhang2022clinical}
\bibinfo{author}{Zhang, L.}, \bibinfo{author}{Chen, D.}, \bibinfo{author}{Song, D.}, \bibinfo{author}{Liu, X.}, \bibinfo{author}{Zhang, Y.}, \bibinfo{author}{Xu, X.}, \bibinfo{author}{Wang, X.}, \bibinfo{year}{2022}.
\newblock \bibinfo{title}{Clinical and translational values of spatial transcriptomics}.
\newblock \bibinfo{journal}{Signal Transduction and Targeted Therapy} \bibinfo{volume}{7}, \bibinfo{pages}{111}.
%Type = Article
\bibitem[{Zhou et~al.(2020)Zhou, Cui, Hu, Zhang, Yang, Liu, Wang, Li and Sun}]{zhou2020graph}
\bibinfo{author}{Zhou, J.}, \bibinfo{author}{Cui, G.}, \bibinfo{author}{Hu, S.}, \bibinfo{author}{Zhang, Z.}, \bibinfo{author}{Yang, C.}, \bibinfo{author}{Liu, Z.}, \bibinfo{author}{Wang, L.}, \bibinfo{author}{Li, C.}, \bibinfo{author}{Sun, M.}, \bibinfo{year}{2020}.
\newblock \bibinfo{title}{Graph neural networks: A review of methods and applications}.
\newblock \bibinfo{journal}{AI open} \bibinfo{volume}{1}, \bibinfo{pages}{57--81}.
%Type = Article
\bibitem[{Zimmermann et~al.(2024)Zimmermann, Vorontsov, Viret, Casson, Zelechowski, Shaikovski, Tenenholtz, Hall, Fuchs, Fusi, Liu and Severson}]{zimmermann2024virchow2}
\bibinfo{author}{Zimmermann, E.}, \bibinfo{author}{Vorontsov, E.}, \bibinfo{author}{Viret, J.}, \bibinfo{author}{Casson, A.}, \bibinfo{author}{Zelechowski, M.}, \bibinfo{author}{Shaikovski, G.}, \bibinfo{author}{Tenenholtz, N.}, \bibinfo{author}{Hall, J.}, \bibinfo{author}{Fuchs, T.}, \bibinfo{author}{Fusi, N.}, \bibinfo{author}{Liu, S.}, \bibinfo{author}{Severson, K.}, \bibinfo{year}{2024}.
\newblock \bibinfo{title}{Virchow2: Scaling self-supervised mixed magnification models in pathology}.
\newblock \bibinfo{journal}{arXiv preprint arXiv:2408.00738} .

\end{thebibliography}
	
	\bio{}
	\endbio
	
	\endbio
	
\end{document}